\documentclass[letterpaper]{article} 

\newif\ifarxivversion
\arxivversiontrue

\ifarxivversion
  \usepackage{aaai2026_arxiv}
\else
  \usepackage{aaai2026}  
\fi

\usepackage{times}  
\usepackage{helvet}  
\usepackage{courier}  
\usepackage[hyphens]{url}  
\usepackage{graphicx} 
\usepackage{natbib}  
\usepackage{caption} 
\usepackage{amsmath}
\usepackage{adjustbox}
\usepackage{xcolor}
\usepackage{textcomp}
\graphicspath{{figures/}}
\usepackage{verbatim}
\usepackage{booktabs}

\ifarxivversion
  \usepackage[hidelinks,breaklinks=true]{hyperref}
  \newcommand{\repositorylink}{%
    \href{https://github.com/Jingxiang-Zhang/tabpfn-multimodal-embeddings}{https://github.com/Jingxiang-Zhang/tabpfn-multimodal-embeddings}%
  }
\else
  \newcommand{\repositorylink}{%
    \url{https://github.com/Jingxiang-Zhang/tabpfn-multimodal-embeddings}%
  }
\fi

\title{TabPFN beyond Tabular Data: Calibration and Accuracy on Multimodal Embeddings}

\author{
    Jingxiang Zhang\textsuperscript{\rm 1}\equalcontrib,
    Lujia Zhong\textsuperscript{\rm 1}\equalcontrib,
    Zijie Zhu\textsuperscript{\rm 1},
    Shuo Huang\textsuperscript{\rm 1},
    Yuang Xu\textsuperscript{\rm 1},
}

\affiliations{
    \textsuperscript{\rm 1}University of Southern California\\
    jzhang92@usc.edu, lujiazho@usc.edu, zzijie@usc.edu, shuohuan@usc.edu, yuangxu@usc.edu
}

\begin{document}
\makeatletter
\def\copyright@on{F}
\makeatother
\maketitle

\begin{abstract}
Few-shot multimodal classification commonly attaches a lightweight head, such as $k$-nearest neighbors, logistic regression, or a linear SVM, to a frozen pretrained encoder. Although computationally efficient, these heads can produce poorly calibrated confidence scores. We ask whether TabPFN can provide reliable confidence estimates on multimodal embeddings without sacrificing predictive accuracy, and under what conditions. We systematically evaluate TabPFN as a zero-gradient head for frozen image, text, and audio encoders. Across 22{,}820 evaluation episodes spanning 14 datasets, 11 encoders, and three modalities, TabPFN achieves the best mean rank among nine classification heads on both negative log-likelihood (NLL) and expected calibration error (ECE). At a representative setting, it reduces NLL by 48--62\% and ECE by 2.1--5.3$\times$ relative to the average of eight baselines while matching or exceeding their average accuracy. This calibration benefit transfers broadly, whereas the accuracy advantage is conditional: it concentrates at moderate-to-high shot counts and low-to-moderate feature dimensions ($k \ge 50$, $d \le 32$), and diminishes when labeled data are scarce, feature dimensions are high, or competing methods approach ceiling accuracy. After backbone adaptation, replacing the trained linear head with TabPFN improves calibration while preserving competitive accuracy, showing that representation adaptation and reliable head choice are complementary. Together, these results identify when TabPFN can serve as a training-free head for calibration-sensitive multimodal classification. To support transparency and reproducibility, we publicly release the source code, experiment configurations, and evaluation scripts in our GitHub repository: \repositorylink.
\end{abstract}

\section{Introduction}
\label{sec:intro}

Attaching a lightweight classification head to a frozen pretrained encoder is a widely used strategy for few-shot multimodal classification~\citep{radford2021learning,bar2024frozen}. The workflow embeds data with a pretrained vision, language, or audio encoder, fits a head such as $k$-nearest neighbors ($k$NN), logistic regression, or linear SVM, and deploys the resulting classifier. Although this gradient-free workflow is computationally efficient, the head must preserve predictive accuracy and provide reliable confidence estimates.

Most classification heads produce poorly calibrated confidence scores. Standard benchmarks rarely expose this issue because they focus solely on accuracy. A classifier is well-calibrated if its stated confidence matches empirical accuracy; that is, a prediction labeled 70\% confident should be correct 70\% of the time. However, these classification heads often fail to satisfy this property~\citep{niculescu2005predicting,silva2023classifier}. In medical imaging triage and clinical decision support, unreliable confidence can be dangerous even when accuracy is high~\citep{pias2025low,rezaeian2026explainability}. Nevertheless, calibration across frozen-encoder and backbone-adaptation pipelines remains underexplored.

Behind this gap lies a fundamental technical challenge: standard classification methods struggle to naturally produce calibrated probabilities. $k$NN and logistic regression can produce miscalibrated probabilities~\citep{niculescu2005predicting,silva2023classifier}, while linear SVMs do not provide probabilities. Linear probing and fine-tuning, including LoRA~\citep{hu2022lora}, may improve accuracy without directly optimizing calibration. Classifiers built on adapted backbones therefore often remain poorly calibrated~\citep{guo2017calibration,he2023preserving,chen2023close}.

To compensate for these deficiencies, post-hoc calibration has become the standard remedy. Platt scaling is commonly applied to SVMs~\citep{platt1999probabilistic}, while temperature scaling~\citep{guo2017calibration}, isotonic regression~\citep{zadrozny2002transforming}, and histogram binning~\citep{berta2024classifier} refine probabilistic predictions. However, these approaches introduce important trade-offs. Platt scaling is limited by its parametric form and may fail to correct more complex miscalibration~\citep{silva2023classifier}. Accuracy-preserving techniques such as temperature scaling require an additional optimization stage and a held-out calibration set. More flexible methods, including isotonic regression and histogram binning, often improve calibration at the expense of predictive accuracy. Moreover, their effectiveness is inconsistent across classifiers and datasets~\citep{manokhin2026classifier}. Thus, no post-hoc strategy consistently provides strong calibration with zero additional training overhead for frozen-encoder pipelines.

TabPFN~\citep{muller2021transformers,hollmann2025accurate} is a promising alternative. It uses tabular in-context learning (ICL): at inference time, it conditions on a table of labeled encoder embeddings and predicts class distributions for test embeddings without updating its parameters. This differs from multimodal ICL with LLMs, where input data are encoded as tokens and inserted into the LLM input. Because TabPFN learns to approximate these distributions during pretraining, calibration is part of its learning objective. Recent studies have applied TabPFN to encoder-derived image, text, and medical-imaging features, showing that it can operate beyond conventional tabular inputs~\citep{walter2026images,dodig2026environmental,neukirch2026foundation}. In this work, we use TabPFN~v2.5~\citep{grinsztajn2025tabpfn} as a training-free classification head applied to the representations of frozen multimodal encoders, which was the latest public release when this study began. TabPFN~v2.6 and v3~\citep{grinsztajn2026tabpfn} were released afterward and are compared post hoc in Ablation~3. Our pipeline has four steps: \emph{encode} $\to$ \emph{PCA-reduce} $\to$ \emph{TabPFN forward pass} $\to$ \emph{class probabilities}. Crucially, this approach requires neither gradient updates nor a held-out calibration set.

TabPFN was trained on synthetic tabular data, and existing applications do not establish whether its calibration transfers across modalities or when it retains competitive accuracy. We therefore ask: Can TabPFN provide reliable confidence estimates on multimodal embeddings without sacrificing predictive accuracy, and under what conditions? To answer this question, we make three connected contributions:

\begin{enumerate}
    \item \textbf{Cross-modal reliability characterization.} We show that TabPFN transfers effectively from tabular pretraining to multimodal embeddings as a zero-gradient head. Across 14 datasets, 11 encoders, and 3 modalities (image, text, and audio), we jointly evaluate predictive performance and calibration against eight alternative classification heads.
    \item \textbf{Mapping the accuracy--reliability boundary.} We identify when calibration is accompanied by an accuracy advantage. The advantage is largest at moderate-to-high shot counts and low-to-moderate feature dimensions, but diminishes when labeled data are scarce, feature dimensions are high, or baselines approach ceiling accuracy.
    \item \textbf{Reliability after backbone adaptation.} We extend the same question to adapted representations. Replacing the trained head with TabPFN substantially improves calibration while preserving competitive accuracy, without additional gradient updates or a held-out calibration set.
\end{enumerate}

\section{Related Work}
\label{sec:related}

\subsubsection{Calibration of Classifiers and Post-Hoc Remedies.}
\citet{niculescu2005predicting} showed that maximum-margin methods produce sigmoid-shaped probability distortions, while \citet{dormann2020calibration} quantified that raw classifier predictions can deviate by up to 0.2 from true probabilities. Post-hoc methods such as Platt scaling~\citep{platt1999probabilistic}, isotonic regression~\citep{zadrozny2002transforming}, and BBQ~\citep{naeini2015obtaining} require a held-out calibration set and additional fitting, and no single method dominates uniformly~\citep{manokhin2026classifier}. TabPFN instead produces probabilities directly from its prior-fitted objective without separate calibration optimization~\citep{muller2021transformers}.

\subsubsection{Frozen-Encoder Classification.}
Training a lightweight head on frozen pretrained encoder representations is widely used for efficient few-shot classification~\citep{radford2021learning,bar2024frozen,sun2019image,abubakar2026lightweight}. Prior work on this paradigm focuses almost exclusively on accuracy, while calibration of the classification head is rarely evaluated~\citep{bar2024frozen,silva2023classifier}. We extend this paradigm by evaluating both calibration and predictive performance when TabPFN is used as a training-free classification head on frozen multimodal embeddings.

\subsubsection{Fine-Tuning and Calibration Failure.}
Fine-tuning pretrained models achieves strong accuracy but often produces poorly calibrated outputs. \citet{guo2017calibration} showed that fine-tuned vision models are overconfident; \citet{he2023preserving} demonstrated that fine-tuning language models degrades pretrained calibration; and \citet{chen2023close} confirmed that confidence increases continuously throughout fine-tuning regardless of correctness. Parameter-efficient methods such as LoRA~\citep{hu2022lora} reduce compute cost but do not resolve the calibration problem. We show that after a backbone has been fine-tuned, replacing its linear head with TabPFN substantially reduces calibration error without requiring any subsequent gradient updates.

\subsubsection{Prior-Fitted Networks and TabPFN.}
Prior-Fitted Networks (PFNs) are models trained to approximate Bayesian inference in-context by repeatedly processing synthetic datasets drawn from a defined prior~\citep{muller2021transformers}. Rather than explicitly targeting the posterior during training, PFNs minimize the prior-data negative log-likelihood of held-out examples. As proven by \citet{muller2021transformers}, this objective is mathematically equivalent to minimizing the expected Kullback-Leibler (KL) divergence between the model's approximation and the true posterior predictive distribution. It therefore provides a principled basis for calibration under the training prior~\citep{muller2025position}, consistent with the view of in-context learning as implicit Bayesian inference~\citep{xie2021explanation}. TabPFN~\citep{hollmann2022tabpfn,hollmann2025accurate} applies this framework to tabular classification. Later versions extend its scale, accuracy, and calibration~\citep{grinsztajn2025tabpfn,grinsztajn2026tabpfn}. \citet{helli2024drift} showed that TabPFN maintains stronger ECE than XGBoost even under temporal distribution shift.

\subsubsection{TabPFN on Non-Tabular Representations.}
Recent studies have begun pairing TabPFN with representations produced by non-tabular encoders. Walter et al.~\citep{walter2026images} encode images with a frozen DINOv3 backbone, apply PCA, and use TabPFN for low-data AI-generated-image detection. Dodig et al.~\citep{dodig2026environmental} evaluate TabPFN on sentence embeddings for ESG sentiment classification, while Neukirch et al.~\citep{neukirch2026foundation} benchmark it among classification heads for CT foundation-model features. Other work learns the interface between multimodal representations and TabPFN. MultiModalPFN introduces modality projectors~\citep{kim2026multimodalpfn}, PromptDx trains an adapter for multimodal Alzheimer's diagnosis~\citep{zhong2026promptdx}, and the TabPFN Text Adapter learns a text-to-TabPFN token projection~\citep{tajjar2026text}. These studies establish task-specific feasibility but do not systematically characterize calibration and accuracy boundaries across image, text, and audio, or evaluate whether a training-free TabPFN head remains useful after backbone adaptation. Our work focuses on this cross-modal reliability question and its operating conditions.

\begin{figure*}[t]
\centering
\includegraphics[width=\textwidth]{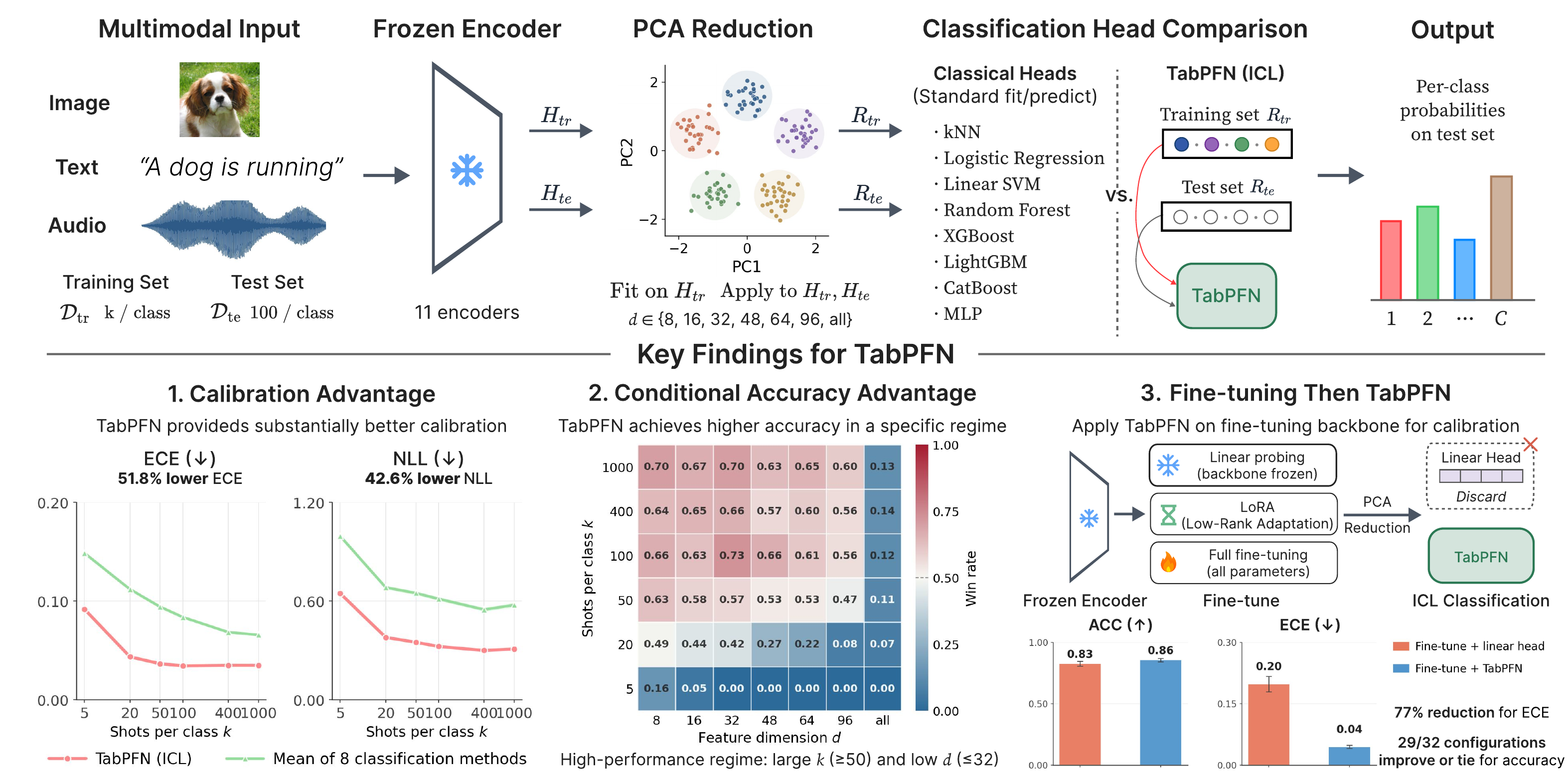}
\caption{%
Overview of the evaluation pipeline and key findings for TabPFN on multimodal embeddings.
\emph{Top (Pipeline):} Multimodal inputs (image, text, audio) are processed through a frozen encoder and reduced via PCA. TabPFN is evaluated as an in-context learning (ICL) classification head and compared against 8 classical methods (e.g., $k$NN, Logistic Regression). 
\emph{Bottom (Key Findings):} (1) Calibration transfers broadly, with TabPFN reducing ECE and NLL relative to the average of eight classical methods. (2) Its accuracy advantage is conditional, concentrating at moderate-to-high shot counts ($k \ge 50$) and low-to-moderate feature dimensions ($d \le 32$). (3) After backbone adaptation, replacing the trained linear head with TabPFN improves calibration while preserving competitive accuracy.%
}
\label{fig:pipeline}
\end{figure*}

\section{Experimental Setup}
\label{sec:setup}

\subsection{Datasets and Encoders}

We evaluate on 14 benchmark datasets spanning three modalities: image, text, and audio (Table~\ref{tab:datasets}). Datasets range from coarse object recognition (CIFAR-10, STL-10) to fine-grained visual categorization (CUB-200-2011, Stanford Dogs), multi-class topic classification (20 Newsgroups), and environmental sound recognition (ESC-10, the 10-class ESC-50 subset), providing a broad spectrum of task difficulties and domain types.

\begin{table}[h]
\centering
\caption{Datasets used in this study, grouped by modality.}
\label{tab:datasets}
\begin{adjustbox}{max width=0.98\columnwidth}
\small
\begin{tabular}{llr}
\hline
\textbf{Dataset} & \textbf{Task} & \textbf{\#Classes} \\
\hline
\multicolumn{3}{l}{\textit{Image}} \\[1pt]
CIFAR-10~\citep{krizhevsky2009learning} & Object recognition & 10 \\
Mini-ImageNet~\citep{vinyals2016matching} & Object recognition & 100 \\
CUB-200-2011~\citep{WahCUB_200_2011} & Fine-grained classification & 200 \\
Stanford Dogs~\citep{khosla2011novel} & Fine-grained classification & 120 \\
STL-10~\citep{coates2011analysis} & Object recognition & 10 \\
\hline
\multicolumn{3}{l}{\textit{Text}} \\[1pt]
20 Newsgroups~\citep{lang1995newsweeder} & Topic classification & 20 \\
AG News~\citep{zhang2015character} & Topic classification & 4 \\
Amazon Reviews~\citep{mcauley2015image} & Sentiment classification & 5 \\
IMDB~\citep{maas2011learning} & Sentiment classification & 2 \\
LLM Emotion~\citep{zhang2025decoding} & Emotion classification & 7 \\
SST-2~\citep{socher2013recursive} & Sentiment classification & 2 \\
\hline
\multicolumn{3}{l}{\textit{Audio}} \\[1pt]
AudioMNIST~\citep{becker2018interpreting} & Digit recognition & 10 \\
Speech Commands~\citep{warden2018speech} & Keyword spotting & 35 \\
ESC-10~\citep{piczak2015esc} & Sound classification & 10 \\
\hline
\end{tabular}
\end{adjustbox}
\end{table}

11 pretrained encoders are used, spanning image, text, and audio modalities (Table~\ref{tab:encoders}). Encoders cover four architecture families, including supervised CNNs (ResNet-50), supervised transformers (ViT-B/16, AST, Whisper), self-supervised transformers (DINOv2-B), and contrastive models (CLIP, CLAP, MiniLM, mpnet, E5), ranging from older supervised baselines to modern, multimodal foundation models. Each modality's encoders are evaluated on all datasets of that modality. All encoders are frozen during Experiments~1 and~2; fine-tuning is compared only in Experiment~3.

\begin{table}[h]
\centering
\caption{Pretrained encoders used in this study, grouped by modality. ``Dim'' means the raw output embedding dimension.}
\label{tab:encoders}
\begin{adjustbox}{max width=\columnwidth}
\small
\begin{tabular}{llr}
\hline
\textbf{Encoder} & \textbf{Pretraining} & \textbf{Dim} \\
\hline
\multicolumn{3}{l}{\textit{Image}} \\[1pt]
ResNet-50~\citep{he2016deep} & Supervised (ImageNet) & 2048 \\
ViT-B/16~\citep{dosovitskiy2020image} & Supervised (ImageNet) & 768 \\
DINOv2-B~\citep{oquab2023dinov2} & Self-supervised & 768 \\
CLIP ViT-B/32~\citep{radford2021learning} & Contrastive (image--text) & 512 \\
\hline
\multicolumn{3}{l}{\textit{Text}} \\[1pt]
MiniLM-L6~\citep{reimers2019sentence} & Sentence contrastive & 384 \\
mpnet-base~\citep{reimers2019sentence} & Sentence contrastive & 768 \\
E5-base~\citep{wang2022text} & Weakly-sup.\ contrastive & 768 \\
CLIP (text)~\citep{radford2021learning} & Contrastive (image--text) & 512 \\
\hline
\multicolumn{3}{l}{\textit{Audio}} \\[1pt]
Whisper-tiny~\citep{radford2023robust} & Supervised (ASR) & 384 \\
AST-AudioSet~\citep{gong2021ast} & Supervised (AudioSet) & 768 \\
CLAP-HTSAT~\citep{wu2023large} & Contrastive (audio--text) & 512 \\
\hline
\end{tabular}
\end{adjustbox}
\end{table}

\subsection{Classification Pipelines}

We evaluate two pipelines: a \emph{frozen-encoder pipeline} (Figure~\ref{fig:pipeline}, top) and a \emph{fine-tuning pipeline} (Figure~\ref{fig:pipeline}, bottom right).

\paragraph{Frozen-encoder pipeline.}
Let $\mathcal{D}_{\mathrm{tr}}$ denote the training set ($k$ labeled samples per class, $C$ classes) and $\mathcal{D}_{\mathrm{te}}$ the held-out test set (100 samples per class). Both pass through the frozen encoder, yielding $H_{\mathrm{tr}}$ and $H_{\mathrm{te}}$. We z-normalize each feature dimension using the mean and standard deviation of $H_{\mathrm{tr}}$ (applied to both splits), then fit PCA on the normalized $H_{\mathrm{tr}}$ and apply the same projection to $H_{\mathrm{te}}$, obtaining $R_{\mathrm{tr}}$ and $R_{\mathrm{te}}$ with feature dimension $d \in \{8, 16, 32, 48, 64, 96, \texttt{all}\}$ columns. Table~\ref{tab:heads} lists the nine classification heads. TabPFN~v2.5~\citep{grinsztajn2025tabpfn} performs ICL: $R_{\mathrm{tr}}$ with its labels is the few-shot context, and $R_{\mathrm{te}}$ is the query in a single forward pass. We fix the v2.5 checkpoint throughout Experiments~1--3 so that all main results refer to one model version; newer TabPFN releases are evaluated separately in Ablation~3. The remaining eight methods are each fitted using the library's standard \texttt{fit} function on $(R_{\mathrm{tr}}, \text{labels}_{\mathrm{tr}})$ and evaluated on $R_{\mathrm{te}}$.

\begin{table}[t]
\centering
\caption{Classification methods included in this study. Primary methods are compared in all experiments. Extended methods are evaluated only in Experiments~1--2.}
\label{tab:heads}
\begin{adjustbox}{max width=\columnwidth}
\footnotesize
\begin{tabular}{ll}
\hline
\textbf{Classification method} & \textbf{Tier} \\
\hline
TabPFN v2.5~\citep{grinsztajn2025tabpfn} & Primary \\
$k$-nearest neighbors~\citep{cover1967nearest} & Primary \\
Logistic regression~\citep{cox1958regression} & Primary \\
Linear SVM~\citep{cortes1995support} & Primary \\
Random forest~\citep{breiman2001random} & Extended \\
XGBoost~\citep{chen2016xgboost} & Extended \\
LightGBM~\citep{ke2017lightgbm} & Extended \\
CatBoost~\citep{prokhorenkova2018catboost} & Extended \\
MLP (two-layer)~\citep{rumelhart1986learning} & Extended \\
\hline
\end{tabular}
\end{adjustbox}
\end{table}

\paragraph{Fine-tuning pipeline.}
The encoder backbone is adapted using three methods: \emph{linear probing} (backbone frozen, only a linear head trained), \emph{LoRA}~\citep{hu2022lora} (low-rank adapters on the encoder's attention layers), and \emph{full fine-tuning} (all encoder parameters updated). For all three methods, a linear head is appended to map the encoder output to $C$ class logits. The active parameters, including the linear head, plus any unfrozen backbone or adapter weights, are then optimized using cross-entropy loss on an 80\% subset of $\mathcal{D}_{\mathrm{tr}}$, with early-stopping validation on the remaining 20\% subset. After adaptation, we evaluate two classification paths for each backbone:

\begin{itemize}
    \item Direct output: The trained linear head is applied to the adapted encoder on $\mathcal{D}_{\mathrm{te}}$, producing logits directly.
    \item TabPFN head: The trained linear head is discarded. We apply PCA with a fixed dimension of $d=64$. This transformation is fit on the adapted encoder's outputs from $\mathcal{D}_{\mathrm{tr}}$ and applied to both $\mathcal{D}_{\mathrm{tr}}$ and $\mathcal{D}_{\mathrm{te}}$. We then run TabPFN in-context on these reduced features. During this step, the adapted encoder remains frozen and requires no further gradient updates.
\end{itemize}

It is worth noting that the path for applying the TabPFN head to a backbone adapted via linear probing is mathematically equivalent to the frozen-encoder baseline. Because linear probing updates only the classification head and leaves the encoder weights unchanged, discarding that trained head exposes the original, unaltered embeddings.

\subsection{Evaluation Metrics}

We compute four metrics on the held-out test set $\mathcal{D}_{\mathrm{te}}$. For probabilistic metrics (NLL, ECE, ASS), each classification head acquires class probabilities via its library-default \texttt{predict\_proba} function. Linear SVM uses \texttt{SVC(kernel="linear", probability=True)}, which maps decision scores to probabilities via an internal Platt scaling procedure~\citep{platt1999probabilistic}. Aside from this internal SVM behavior, no external post-hoc calibration is applied in the main evaluation grid; we evaluate post-hoc temperature scaling across all nine classification heads in Appendix~I. Let $N$ be the number of test samples, $x_i$ the input, $y_i$ the true label, and $\hat{p}(y|x)$ the predicted probability distribution. The four metrics are defined as follows:

\begin{itemize}
    \item Accuracy: Top-1 classification accuracy.
    
    \item Negative log-likelihood (NLL): The multiclass log-loss on the predicted probability of the true class. Lower values indicate better probabilistic predictions:
    $$ \text{NLL} = -\frac{1}{N}\sum_{i=1}^N \log \hat{p}(y_i | x_i) $$
    
    \item Expected calibration error (ECE): Measures the alignment between predictive confidence and empirical accuracy~\citep{guo2017calibration}. We partition the maximum predicted probabilities into $M=15$ uniform-width bins $B_m$ over $[0,1]$. Lower values indicate better calibration:
    $$ \text{ECE} = \sum_{m=1}^{M} \frac{|B_m|}{N} \left| \text{acc}(B_m) - \text{conf}(B_m) \right| $$
    where $\text{acc}(B_m)$ is the empirical accuracy in bin $m$ and $\text{conf}(B_m)$ is the mean maximum predicted probability among samples in bin $m$.
    
    \item ASS (Average Prediction-Set Size): A metric of selective prediction computed via split conformal Adaptive Prediction Sets (APS)~\citep{romano2020classification}. The goal of APS is to output a set of classes $\mathcal{C}(x_i)$ that contains the true label with a guaranteed probability. We target 90\% coverage, corresponding to a significance level $\alpha=0.1$. To compute this, the test set $\mathcal{D}_{\mathrm{te}}$ is randomly split 50/50 into a calibration subset $\mathcal{D}_{\mathrm{cal}}$ and an evaluation subset $\mathcal{D}_{\mathrm{eval}}$. On each calibration point $(x_i, y_i)$, we compute a nonconformity score $s_i$ as the cumulative probability of all classes ranked above the true label: $s_i = \sum_{j:\,\hat{p}(j|x_i)\geq \hat{p}(y_i|x_i)} \hat{p}(j|x_i)$. We set $\hat{q}$ to the $(1-\alpha)(1+1/|\mathcal{D}_\mathrm{cal}|)$ quantile of $\{s_i\}$, which guarantees the desired $(1-\alpha)=90\%$ coverage on held-out data. For each evaluation point $x_i \in \mathcal{D}_{\mathrm{eval}}$, classes are added in decreasing probability order until their cumulative sum meets or exceeds $\hat{q}$. ASS~\citep{sadinle2019least} measures the average set size:
    $$ \text{ASS} = \frac{1}{|\mathcal{D}_{\mathrm{eval}}|} \sum_{i \in \mathcal{D}_{\mathrm{eval}}} |\mathcal{C}(x_i)| $$
    Lower ASS indicates tighter prediction sets under the same 90\% coverage guarantee.

\end{itemize}

Throughout our experiments, the term baseline refers to any of the non-TabPFN classification heads evaluated in our study (e.g., $k$NN, logistic regression, linear SVM, and tree-based ensembles). When computing win-rates or relative improvements, TabPFN is compared against the maximum performance achieved by any of these baseline methods in that specific setting.

\subsection{Evaluation Grid}
We evaluate a full grid across all datasets, encoders, and 5 independent random seeds, totaling 22{,}820 evaluation episodes. Within each episode, all classification heads are evaluated on the exact same train/test split. The grid parameters are:

\begin{itemize}
    \item Shots per class ($k$-shot): $k \in \{5, 20, 50, 100, 400, 1000\}$
    \item Number of classes: $C \in \{2, 5, 10\}$
    \item Feature dimension: $d \in \{8, 16, 32, 48, 64, 96, \texttt{all}\}$
\end{itemize}

To construct each evaluation episode, we aim to sample $C$ classes and partition the data into $k$ training examples and 100 held-out test examples per class. We enforce two data-sufficiency filters to ensure reliable evaluation. First, configurations requiring more classes than a dataset contains (e.g., evaluating $C=10$ on a binary dataset) are globally excluded. Second, if any sampled class contains fewer than $k + 10$ total examples, the specific episode is discarded to prevent statistically unstable test sets. For valid episodes where a class has at least $k + 10$ but fewer than $k + 100$ examples, we maintain exactly $k$ training examples and allocate the remainder to the test set.

To prevent rank-deficient PCA projections, the final feature dimension is strictly bounded by $\min(d,\, kC - 1,\, d_{\mathrm{enc}},\, 2000)$, where $d_{\mathrm{enc}}$ is the raw encoder dimension and 2000 is TabPFN v2.5's maximum supported feature capacity. For the $d=\texttt{all}$ configuration, we omit the target $d$ and simply extract this maximum allowable dimension.

Due to the computational cost of gradient updates, fine-tuning evaluations utilize a targeted subset of this grid, detailed in section Experiment 3.

\section{Experiments}
\label{sec:experiments}

Three questions structure our empirical evaluation. First, does TabPFN improve calibration on frozen multimodal embeddings? Second, when is this improvement achieved without an accuracy penalty? Finally, can TabPFN complement common backbone-adaptation pipelines?

\subsection{Experiment 1: Does TabPFN Improve Calibration?}
\label{sec:exp1}

\begin{figure*}[t]
\centering
\includegraphics[width=\textwidth]{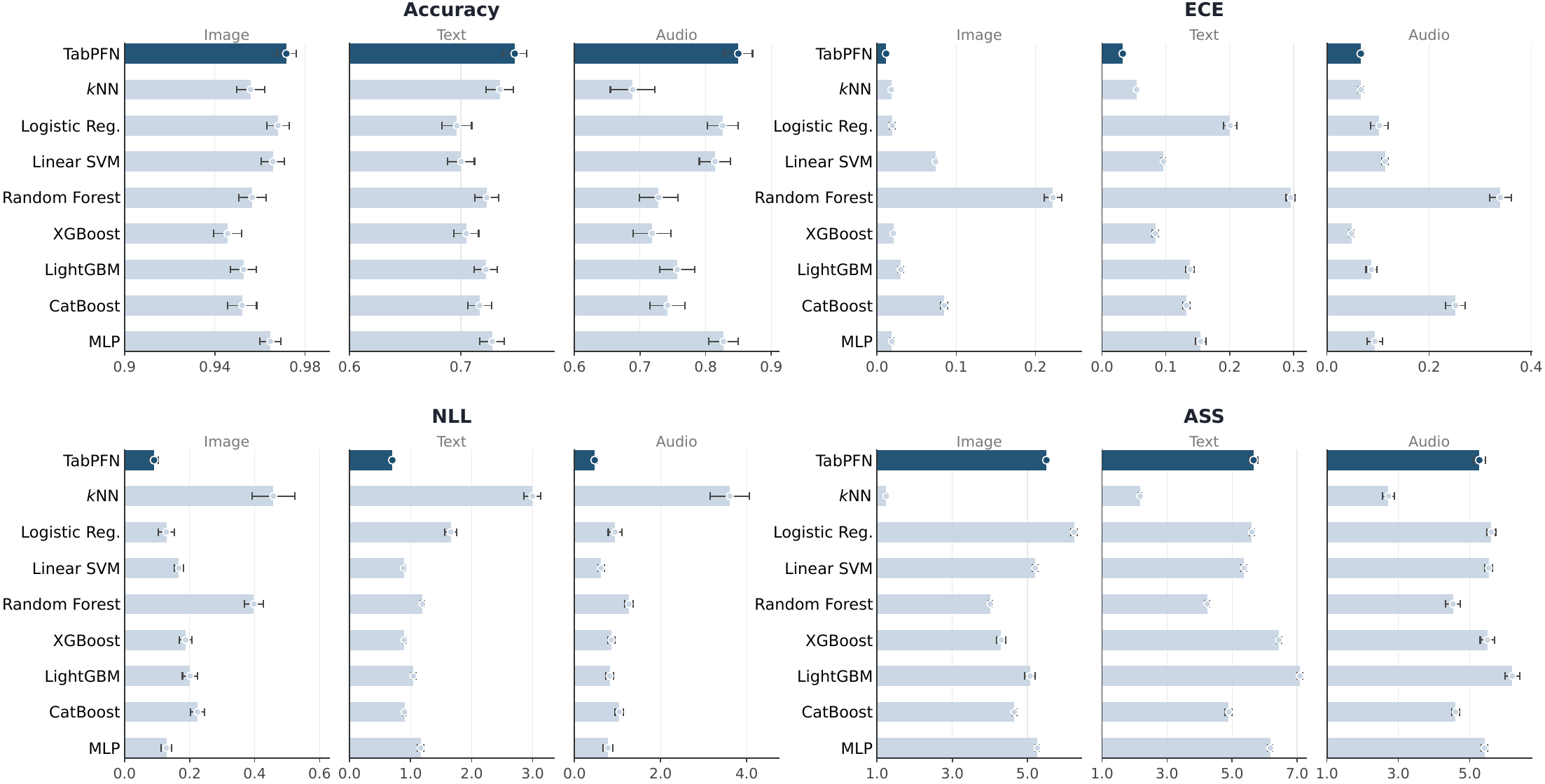}
\caption{All four metrics at the canonical setting ($C=10$, $k=100$, $d=96$). Mean $\pm$ SEM (standard error of the mean) aggregated over all datasets, encoders, and seeds per modality. TabPFN (dark blue) achieves 48--62\% lower NLL (bottom left) and 2.1--5.3$\times$ lower ECE (top right) than the average of all baselines, while surpassing their average accuracy on every modality (top left). On ASS (bottom right), $k$NN's peaked distributions yield the tightest prediction sets.}
\label{fig:exp1}
\end{figure*}

Experiment~1 asks whether TabPFN improves calibration on frozen multimodal embeddings. We first aggregate ranks over all 22{,}820 evaluation episodes (Table~\ref{tab:exp1_rank}) to summarize overall performance, then use a per-modality breakdown at a canonical setting (Figure~\ref{fig:exp1}) to show the practical gap. We also report accuracy to establish predictive usefulness. Experiment~2 directly tests when the calibration improvement comes without an accuracy penalty.

Across all valid evaluation episodes, TabPFN achieves the best mean rank on both calibration metrics: 2.12 $\pm$ 1.45 on NLL and 2.93 $\pm$ 2.00 on ECE (Table~\ref{tab:exp1_rank}). The low standard deviation of these ranks indicates that this calibration advantage is highly consistent on average across the evaluation grid. On accuracy, logistic regression slightly outperforms TabPFN (3.20 $\pm$ 1.93 vs.\ 3.27 $\pm$ 2.01). However, because logistic regression ranks lower on both NLL and ECE, this demonstrates that strong accuracy does not inherently guarantee well-calibrated probabilities. Tree-based methods rank poorly on both accuracy and calibration.

Figure~\ref{fig:exp1} shows the per-modality results at a canonical high-shot setting ($C=10$, $k=100$, $d=96$). We use shot per class $k=100$ as a representative setting with sufficient in-context examples, and $d=96$ because using all raw encoder dimensions ($d=\texttt{all}$) degrades TabPFN's calibration without improving other metrics. At this setting, TabPFN reduces NLL by 48--62\% and achieves an ECE 2.1--5.3$\times$ lower than the average of all baselines, while outperforming their average accuracy on every modality (+1.4 to +8.7 percentage points). This pattern is consistent with TabPFN's Bayesian training objective~\citep{muller2021transformers}: its \texttt{predict\_proba} outputs are designed to approximate posterior predictive distributions, whereas classical heads return confidence scores that can deviate from empirical frequencies. Furthermore, per-head reliability diagrams (Appendix~A, Figure~\ref{fig:appendix1_reliability}) confirm this behavior: linear SVM deviates substantially from the ideal identity line, while TabPFN's curve tracks it better across all three modalities. red To test whether post-hoc recalibration closes this gap, Appendix~I (Figure~\ref{fig:appendix_ts_baselines} and Tables~\ref{tab:appendix_ts_all}--\ref{tab:appendix_ts_paired}) applies 5-fold cross-validated temperature scaling (TS) to all nine heads at $k \in \{100, 400, 1000\}$. While TS reduces baseline ECE to the level of uncalibrated TabPFN, uncalibrated TabPFN still achieves lower NLL and higher accuracy than every TS-calibrated baseline, and TabPFN\,+\,TS achieves the lowest ECE and NLL overall.

\begin{table}[t]
\centering
\caption{Mean rank across all 22{,}820 evaluation episodes (dataset, encoder, seed, $k$, $d$, $C$). Lower is better (rank~1 = best); bold indicates the best mean rank per column.}
\label{tab:exp1_rank}
\begin{adjustbox}{max width=\columnwidth}
\footnotesize
\begin{tabular}{lcccc}
\hline
\textbf{Head} & \textbf{Acc Rank} & \textbf{NLL Rank} & \textbf{ECE Rank} & \textbf{ASS Rank} \\
\hline
TabPFN          & 3.27 $\pm$ 2.01          & \textbf{2.12 $\pm$ 1.45} & \textbf{2.93 $\pm$ 2.00} & 5.51 $\pm$ 1.93          \\
$k$NN           & 5.02 $\pm$ 2.50          & 6.52 $\pm$ 3.02          & 3.79 $\pm$ 2.03          & \textbf{1.18 $\pm$ 0.65} \\
Logistic Reg.   & \textbf{3.20 $\pm$ 1.93} & 3.33 $\pm$ 2.32          & 3.96 $\pm$ 2.26          & 5.99 $\pm$ 1.95          \\
Linear SVM      & 3.96 $\pm$ 2.09          & 4.63 $\pm$ 2.30          & 5.97 $\pm$ 2.40          & 5.31 $\pm$ 2.03          \\
\hline
Random Forest   & 5.71 $\pm$ 1.93          & 6.71 $\pm$ 2.11          & 7.34 $\pm$ 2.44          & 4.39 $\pm$ 2.01          \\
XGBoost         & 7.05 $\pm$ 2.03          & 5.75 $\pm$ 1.66          & 5.16 $\pm$ 1.96          & 5.21 $\pm$ 2.24          \\
LightGBM        & 6.67 $\pm$ 2.06          & 5.94 $\pm$ 2.06          & 4.58 $\pm$ 2.52          & 6.23 $\pm$ 2.15          \\
CatBoost        & 5.53 $\pm$ 2.00          & 5.22 $\pm$ 1.80          & 5.97 $\pm$ 2.21          & 5.18 $\pm$ 1.66          \\
MLP             & 4.60 $\pm$ 2.14          & 4.79 $\pm$ 2.33          & 5.31 $\pm$ 2.36          & 5.99 $\pm$ 1.95          \\
\hline
\end{tabular}
\end{adjustbox}
\end{table}

\begin{figure*}[t]
\centering
\includegraphics[width=\textwidth]{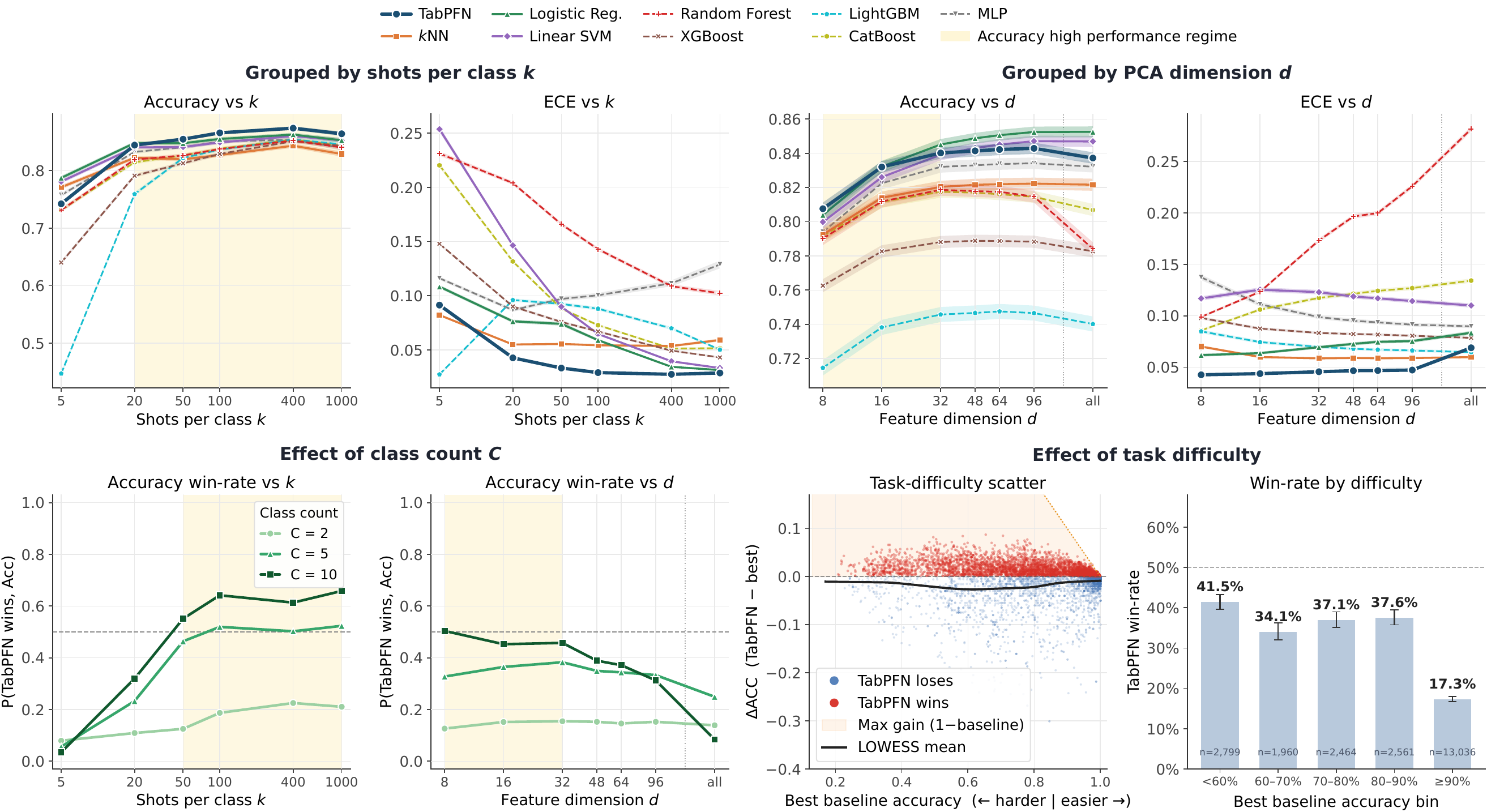}
\caption{%
Calibration transfers broadly, whereas TabPFN's accuracy advantage is conditional.
All panels compare TabPFN against the best of all eight competing heads (best-of-8).
\emph{Top row:} mean $\pm$ SEM of all nine heads versus shots per class $k$ (cols~1--2) and feature dimension $d$ (cols~3--4), aggregated over all datasets, encoders, seeds, and $C\in\{2,5,10\}$ (shaded regions indicate $\pm$1 SEM); the yellow-shaded band marks where TabPFN's accuracy gains concentrate.
\emph{Bottom row:} accuracy win-rate $P(\text{TabPFN}>\text{best-of-8})$ by class count $C$ (cols~1--2), and $\Delta$ACC versus best-baseline accuracy with difficulty-binned win-rates (cols~3--4).%
}
\label{fig:exp2}
\end{figure*}

TabPFN's calibration advantage does not extend to ASS. While $k$NN consistently achieves the tightest prediction sets (mean rank 1.18), TabPFN falls into the lower half of the evaluated methods on this metric (mean rank 5.51). We analyze this conformal set-size gap in the Ablation section.

\subsection{Experiment 2: When Does Calibration Come Without an Accuracy Penalty?}
\label{sec:exp2}

Experiment~1 established that TabPFN broadly improves calibration on frozen embeddings. Experiment~2 tests whether this reliability comes at an accuracy cost and maps the conditions under which it does not. We vary shot per class $k$, feature dimension $d$, class count $C$, and task difficulty. Throughout, we compare TabPFN against the best of all eight competing heads. Figure~\ref{fig:exp2} presents the four main findings; stratified means for all nine heads at $C=10$ are in Appendix~D (Tables~\ref{tab:exp2_shots_stratified} and~\ref{tab:exp2_pca_stratified}), and the complete $(k\times d)$ win-rate breakdown and remaining metrics are in Appendices~B, C, and E (Figures~\ref{fig:exp2_appendix_nll_ass}--\ref{fig:exp2_appendix_heatmaps}).

The top row of Figure~\ref{fig:exp2} reveals an asymmetric boundary: calibration remains robust across most operating conditions, whereas the accuracy advantage is conditional. On ECE, TabPFN achieves lower error than every baseline across all $k \geq 20$ and all $d \leq 96$. NLL shows an even stronger advantage, with TabPFN remaining below every baseline at all $k$ and all $d$, including $d=\texttt{all}$ (Appendix Figure~\ref{fig:exp2_appendix_nll_ass}). Accuracy follows a different pattern. At $k=5$, TabPFN trails logistic regression (0.742 vs.\ 0.787), linear SVM, $k$NN, and MLP, suggesting that five examples may provide insufficient context. Starting from $k=50$, TabPFN leads or ties the field, holding the highest mean accuracy across $k=100$ to $k=1000$ and peaking at 0.874. Along the feature-dimension axis, the advantage concentrates in the yellow $d \in [8, 32]$ band. TabPFN leads in mean accuracy at $d \in [8, 16]$, while logistic regression marginally overtakes it at $d=32$. However, $P(\text{TabPFN} > \text{best-of-8})$ remains stable across $d \in [8, 32]$ before declining at $d \geq 48$, so the small mean difference at $d=32$ does not mark a regime change.

Class count and task difficulty further shape this boundary. Splitting the win-rate curves by class count (Figure~\ref{fig:exp2}, bottom row) gives a clear ordering of $C=10 > C=5 > C=2$ for $d \le 64$. This hierarchy breaks down at $d \ge 96$, consistent with feature-capacity constraints. Binary tasks never cross the 50\% win-rate threshold because the best-of-8 baseline approaches perfect accuracy, leaving little room for improvement. When the best baseline achieves $\ge 90\%$ accuracy (57\% of all episodes), TabPFN's win rate drops to 17.3\%, compared with approximately 37\% on harder episodes. This pattern suggests that additional classes provide more multi-class structure for TabPFN's in-context mechanism to exploit. For $C=5$, the win rate reaches and plateaus near 50\% for $k \ge 100$, and for $C=10$, it exceeds 50\% starting at $k=50$ and peaks at 0.64--0.66.

Appendix Figure~\ref{fig:exp2_appendix_difficulty} further quantifies the ceiling effect. From hard to easy tasks, the standard deviation of $\Delta$ACC drops from 0.054 to 0.025, and the 90th-percentile gain drops from +0.032 to +0.005. This supports a ceiling explanation: strong baselines leave little room for additional accuracy gains. The full $(k\times d)$ maps in Appendix Figure~\ref{fig:exp2_appendix_heatmaps} show the structural counterpart. The winning region expands from nearly empty at $C=2$ to a broad mid-$k$ plateau at $C=10$, while $d=\texttt{all}$ consistently has near-zero win rates, consistent with raw dimensions exceeding TabPFN's effective capacity. Because smaller datasets do not have enough examples for large $k$ and small $k$ caps the maximum PCA dimension, Appendix~E (Figures~\ref{fig:appendix_fixed_grid_winrate}--\ref{fig:appendix_fixed_grid_cell_means}) re-evaluates $C=10$ on fixed groups of dataset--encoder pairs, confirming that all shot and dimension trends still hold. Together, these results locate where calibration and accuracy gains coexist: moderate-to-high shots, low-to-moderate dimensions, multiclass tasks, and sufficient room above the baseline.

\subsection{Experiment 3: Can TabPFN Complement Backbone Adaptation?}
\label{sec:exp3}

\begin{figure*}[t]
\centering
\includegraphics[width=\textwidth]{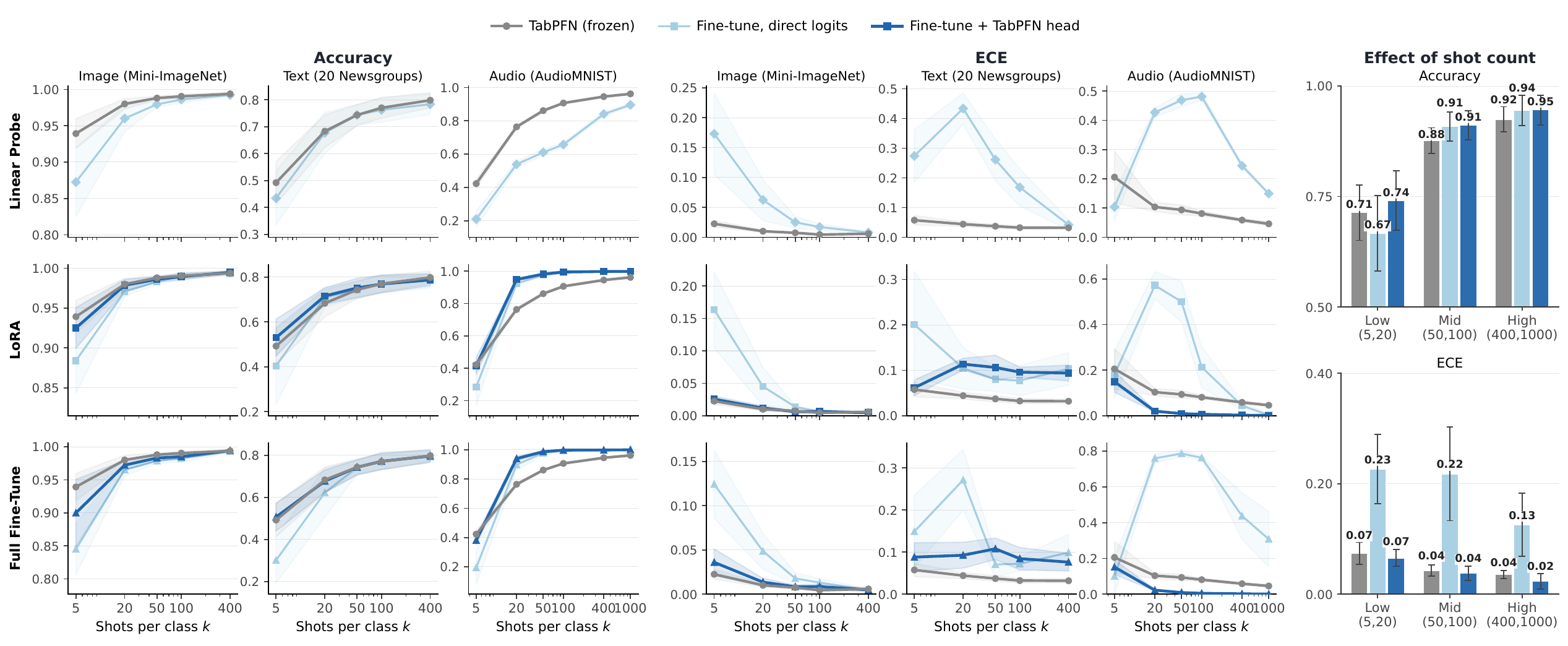}
\caption{%
Accuracy and calibration after backbone adaptation across shot counts.
\emph{Left block}: accuracy versus shots per class $k$; \emph{middle block}: ECE versus shots per class $k$; \emph{right block}: aggregate effect of shots per class count. Rows show the fine-tuning method (Linear Probe, LoRA, Full Fine-Tune), and columns in the line-plot blocks show modality and dataset (Image / Mini-ImageNet, Text / 20 Newsgroups, Audio / AudioMNIST). Grey = TabPFN on frozen encoder (reference); light blue = fine-tune with direct linear head; dark blue = fine-tune $+$ TabPFN head. Shading: mean $\pm$1 std over 5 seeds. The right bar panels aggregate LoRA and Full Fine-Tune configurations into low-shot ($k{\in}\{5,20\}$), mid-shot ($k{\in}\{50,100\}$), and high-shot ($k{\in}\{400,1000\}$) buckets, with bars showing mean $\pm$ SEM across configurations.
}
\label{fig:exp3}
\end{figure*}

Experiments~1 and~2 established the accuracy--reliability boundary for frozen embeddings. Experiment~3 asks whether TabPFN can improve prediction reliability while preserving the accuracy gains from backbone adaptation. We compare three adaptation strategies (linear probing, LoRA~\citep{hu2022lora}, and full fine-tuning) on one representative encoder and dataset per modality (DINOv2-B / Mini-ImageNet, E5-base / 20 Newsgroups, CLAP-HTSAT / AudioMNIST), at a fixed $C=10$, $d=64$, over $k \in \{5,20,50,100,400\}$ ($k=1000$ for audio), with 5 seeds. Each adapted backbone is evaluated through its trained linear head (\emph{direct}) and an alternative path (\emph{+TabPFN}) that discards this head, reduces the adapted features via PCA, and applies TabPFN in-context. This design separates the contribution of representation adaptation from that of head choice. Figure~\ref{fig:exp3} reports accuracy and ECE; NLL and ASS are in Appendix~F (Figure~\ref{fig:exp3_appendix_nll_ass}).

The accuracy panels of Figure~\ref{fig:exp3} first show when representation adaptation is useful and when it overfits. Linear probing trains only a linear head on frozen features and overfits when labels are limited: at $k=5$ (50 total training examples given $C=10$), it trails frozen TabPFN by 6.7 percentage points on Mini-ImageNet ($0.872$ vs.\ $0.939$) and 21.3 percentage points on AudioMNIST ($0.210$ vs.\ $0.423$). As $k$ grows, the head is fit on enough data to close the gap, nearly matching frozen TabPFN on the two well-matched encoders by $k=400$ (image $0.992$ vs.\ $0.994$; text $0.783$ vs.\ $0.798$). Audio is the exception: the CLAP-HTSAT backbone was pretrained on general audio--text pairs (LAION-Audio-630K and AudioSet)~\citep{wu2023large}, a distribution far from spoken digits. Consequently, a frozen linear head cannot adapt the representation enough and still trails frozen TabPFN by 6.6 percentage points at $k=1000$ ($0.896$ vs.\ $0.962$).

Full fine-tuning updates the entire backbone and overfits even more aggressively at very low sample sizes. At $k=5$, it falls below even linear probing on all three datasets ($0.301$ vs.\ $0.435$ on 20 Newsgroups, $0.194$ vs.\ $0.210$ on AudioMNIST). Once a moderate amount of data is available, updating the backbone yields the largest gains on the mismatched audio encoder: at $k=20$, full fine-tuning reaches $0.898$ on AudioMNIST, moving above both frozen TabPFN ($0.764$) and linear probing ($0.539$). LoRA combines the strengths of both. Its low-rank adapters mitigate overfitting at small $k$, outperforming full fine-tuning ($0.285$ vs.\ $0.194$ on AudioMNIST, $0.404$ vs.\ $0.301$ on 20 Newsgroups at $k=5$), while adapting the backbone as effectively as full fine-tuning at larger $k$ (AudioMNIST: $0.982$ vs.\ $0.987$ at $k=50$; $0.995$ vs.\ $0.997$ at $k=100$).

TabPFN complements this representation adaptation. Replacing a fine-tuned linear head with TabPFN improves or ties accuracy in 29 of the 32 configurations. The magnitude of this gain remains difficulty-dependent. Using direct-head accuracy as a difficulty proxy, the mean gain is 14.1 percentage points below $70\%$ accuracy (hard tasks), 1.6 percentage points in the $70\%$--$90\%$ range, and 0.4 percentage points above $90\%$ (easy tasks). The effect is most pronounced on the mismatched audio backbone: LoRA first increases AudioMNIST accuracy above the frozen-TabPFN reference ($0.925$ vs.\ $0.764$ at $k=20$), and applying TabPFN to these adapted features increases it further to $0.948$. Because linear probing leaves the backbone unchanged, linear probe\,$+$\,TabPFN is equivalent to the frozen-encoder baseline and is omitted from the accuracy panels in Figure~\ref{fig:exp3}.

Head choice separates more clearly on calibration. The middle block of Figure~\ref{fig:exp3} shows that frozen TabPFN and fine-tune\,$+$\,TabPFN are better calibrated than direct fine-tuned heads almost everywhere. On audio, the direct heads are severely miscalibrated in the mid-shot range (full fine-tuning peaks at an ECE of $0.79$ on AudioMNIST at $k=50$, because early stopping on validation accuracy prevents further confidence adjustment; see Appendix~F, Figure~\ref{fig:appendix_audiomnist_rel}), while both TabPFN approaches stay an order of magnitude lower. Fine-tuning aligns the mismatched CLAP-HTSAT backbone to the target task, and applying TabPFN to these adapted features yields an ECE of $0.001$--$0.020$ for $k{\geq}20$, substantially outperforming the frozen-TabPFN baseline ($0.046$--$0.104$). Text shows a different trend: although fine-tune\,$+$\,TabPFN remains better calibrated than the direct head on 20 Newsgroups, its ECE ($0.06$--$0.11$) is higher than that of frozen TabPFN ($0.03$--$0.06$). NLL follows the same pattern across all three modalities (Appendix~F, Figure~\ref{fig:exp3_appendix_nll_ass}).

The right-hand panels of Figure~\ref{fig:exp3} summarize this complementarity over low-shot ($k \in \{5,20\}$), mid-shot ($k \in \{50,100\}$), and high-shot ($k \in \{400,1000\}$) regimes, aggregating LoRA and full fine-tuning. Linear probing is excluded because it does not update the backbone. At low shots, direct fine-tuning yields 0.67 accuracy, compared with 0.71 for frozen TabPFN and 0.74 for fine-tune\,$+$\,TabPFN. At mid shots, both adapted paths reach approximately 0.91 versus 0.88 for frozen TabPFN. At high shots, the direct and fine-tune\,$+$\,TabPFN paths reach 0.94 and 0.95, respectively, versus 0.92 for the frozen baseline.

Calibration separates more clearly in the aggregated results. Direct fine-tuning yields ECE from 0.13 to 0.23. Replacing the linear head with TabPFN matches the frozen baseline at low and mid shots (ECE 0.07 and 0.04) and surpasses it at high shots (ECE 0.02 vs. 0.04). Appendix Figure~\ref{fig:exp3_appendix_nll_ass} confirms the same trend for NLL. Fine-tune\,$+$\,TabPFN has the lowest mean log-loss in all three shot regimes, although TabPFN-based heads produce larger prediction sets than direct fine-tuned heads (5.06--6.23 vs.\ 3.79--5.95). Together, these results show that backbone adaptation and TabPFN address different parts of the pipeline: adaptation improves task-specific representations, while TabPFN provides a more reliable classification head. In Appendix~F (Figures~\ref{fig:appendix_exp3_ts_acc_ece}--\ref{fig:appendix_exp3_ts_nll_ass}), we further evaluate these methods using temperature scaling.

\section{Ablation Studies}
\label{sec:ablations}

We report three ablations with distinct designs. Ablation~1 tests whether the calibration transfer and accuracy--reliability boundary from Experiments~1--2 are robust to preprocessing and label noise, using a full $(k\times d)$ win-rate matrix on one representative dataset per modality. Ablation~2 diagnoses the discrepancy between TabPFN's strong calibration and its poor ASS performance from Experiment~1 via Top-$k$ calibration at the canonical $C=10$, $k=100$, $d=96$ setting. Ablation~3 compares seven ICL foundation models against the classical heads from Experiment~1 at the same canonical setting. This comparison investigates whether strong calibration is unique to TabPFN or a broader property of ICL tabular heads.

\subsection{Ablation 1: Preprocessing Methods and Label-Noise Sensitivity}
\label{sec:ablation_robustness}

\begin{figure}[t]
\centering
\includegraphics[width=\columnwidth]{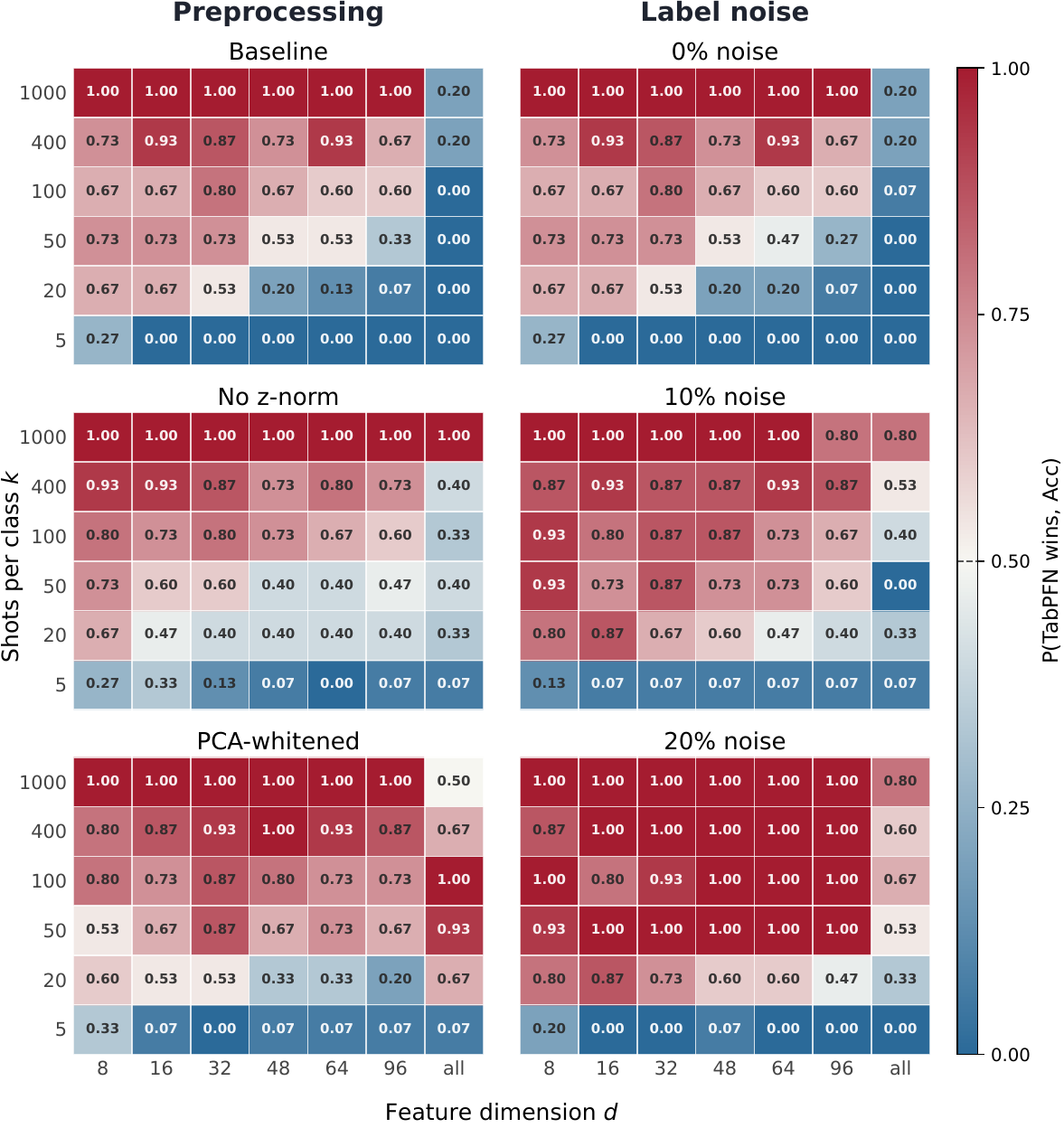}
\caption{%
Accuracy win-rate at each $(k, d)$, aggregated over Mini-ImageNet, 20 Newsgroups, and AudioMNIST ($C=10$, 5 seeds). \emph{Left:} three preprocessing variants. \emph{Right:} three label-noise levels (training set only).%
}
\label{fig:ablation}
\end{figure}

To verify that the accuracy-favorable region does not rely on a specific preprocessing pipeline or perfectly clean training labels, we re-run the full $(k\times d)$ grid under alternative settings. Holding $C=10$ fixed, we use one representative encoder and dataset per modality (DINOv2-B / Mini-ImageNet, E5-base / 20 Newsgroups, CLAP-HTSAT / AudioMNIST, 5 seeds each) and recompute TabPFN's win-rate under three preprocessing variants and three symmetric label-noise levels (applied to the training set only). Figure~\ref{fig:ablation} reports the resulting heatmaps.

The win-rate pattern is consistent across all three preprocessing variants (Figure~\ref{fig:ablation}, left column). The high-win-rate regime for $k \geq 20$ at low-to-mid feature dimensions $d$ appears uniformly. At the maximum available PCA dimension ($d=\texttt{all}$), whitening substantially increases TabPFN's win-rate without materially changing its absolute accuracy, indicating that TabPFN is less sensitive than the competing heads to the amplification of low-variance components. Standard z-normalization is adopted as the default preprocessing method throughout our main experiments. In Appendix~J (Tables~\ref{tab:appendix_dim_reduction}--\ref{tab:appendix_pca_cost}), we also compare PCA against non-linear and supervised dimensionality reduction methods.

We inject symmetric uniform label noise into the training set at $10\%$ and $20\%$, leaving the test labels uncorrupted. As noise increases, TabPFN's win-rate rises (Figure~\ref{fig:ablation}, right column): the high-win-rate regime expands from the $0\%$ panel to the $20\%$ panel, indicating that TabPFN is more robust to label corruption than the classical baselines. Notably, the $d=\texttt{all}$ column also improves under noise: at $10\%$ noise, TabPFN's win-rate at $d=\texttt{all}$ exceeds $0.5$ for $k \geq 400$, and at $20\%$ noise, this win-rate remains above $0.5$ for all $k \geq 50$ ($0.53$--$0.80$).

\subsection{Ablation 2: Top-$k$ Calibration and Selective Prediction}
\label{sec:ablation_ass}

Experiment~1 shows that while TabPFN ranks first on calibration metrics like ECE and NLL, it ranks last on ASS. A natural explanation is that TabPFN miscalibrates probabilities beyond the top-ranked class: if the model systematically assigns miscalibrated probability mass to ranks 2--5, APS may require long prefixes to reach the coverage threshold. We fix the canonical setting ($C=10$, $k=100$, $d=96$) and evaluate all four primary heads. For each classification head, we compute Top-$k$ ECE for $k\in\{1,\ldots,5\}$: samples are binned by their cumulative top-$k$ confidence $p_{\mathrm{sum}}=\sum_{j\le k}\hat{p}_{(j)}$, and the empirical accuracy of each bin is simply the frequency that the true label falls within those top $k$ predictions. At $k=1$, this reduces to standard ECE; for $k>1$, it directly evaluates the calibration of the probabilities beyond the argmax.

\begin{figure}[t]
\centering
\includegraphics[width=1.0\columnwidth]{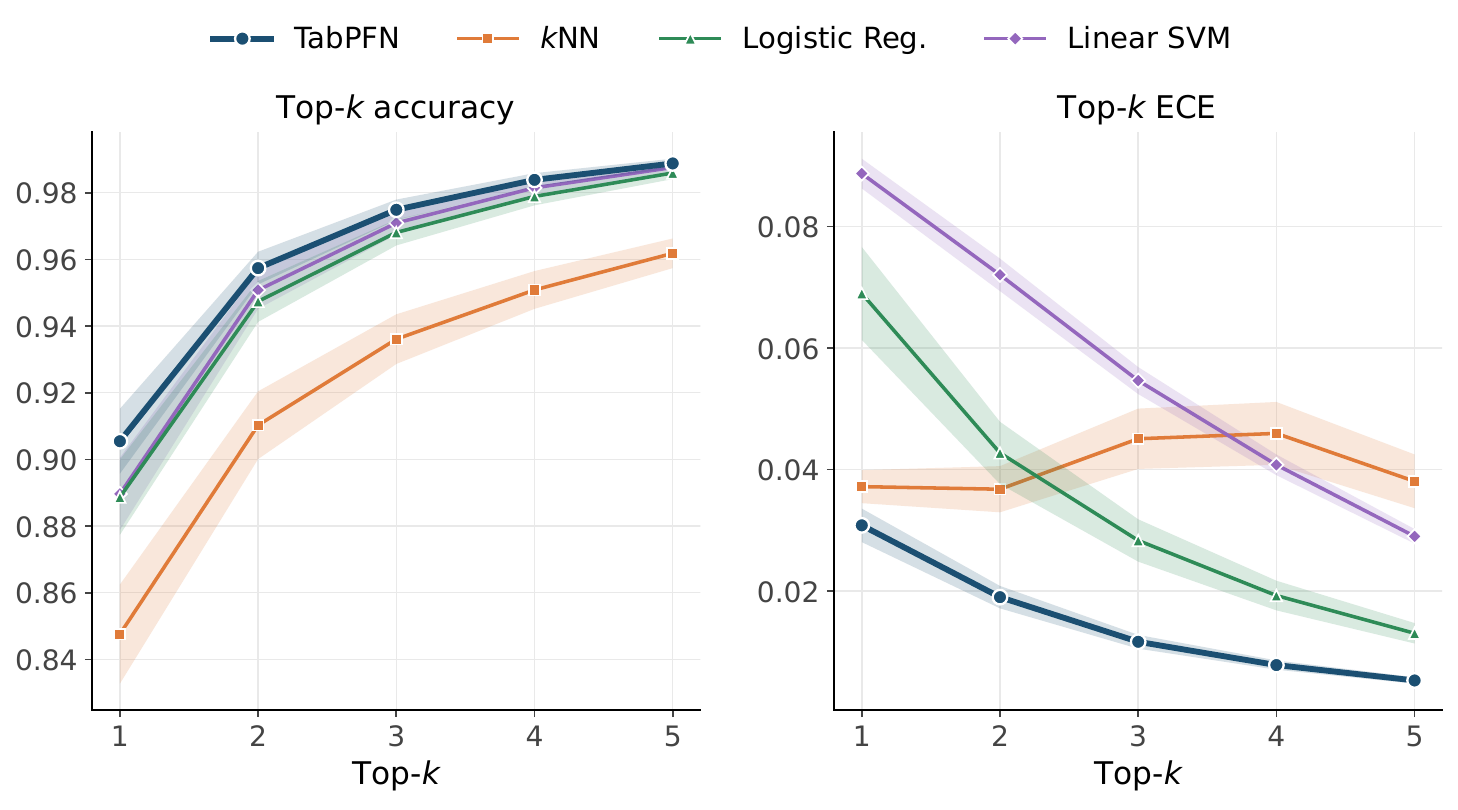}
\caption{%
Top-$k$ calibration at the canonical setting ($C=10$, $k=100$, $d=96$), mean $\pm$1 SEM over 560 ablation episodes. \emph{Left:} Top-$k$ accuracy. \emph{Right:} Top-$k$ ECE.%
}
\label{fig:ablation_topkp_metrics}
\end{figure}

Figure~\ref{fig:ablation_topkp_metrics} and Table~\ref{tab:ablation_ass} present the aggregated results. TabPFN achieves the lowest Top-$k$ ECE for every $k$ from 1 to 5, and its calibration advantage over $k$NN widens as $k$ increases (Top-5 ECE: 0.005 vs. 0.038). After ranking the heads within each evaluation episode and averaging these ranks across episodes, TabPFN achieves a mean Top-5 ECE rank of 1.48, compared with 3.16 for $k$NN. (Appendix G, Table~\ref{tab:ablation_topkp_ranks}). Miscalibration of the probability mass beyond the top-1 class therefore does not explain TabPFN’s larger conformal prediction sets.

The difference in ASS performance stems directly from the shape of the predicted posteriors. For $k$NN, which uses distance-weighted voting over the $5$ nearest neighbors in our pipeline, classes absent from the neighborhood receive exactly zero probability, and the prediction frequently collapses to a point mass. When all neighbors agree, the model outputs $\hat{p}(\hat{y}|x)=1.0$, assigning zero probability to all other classes. Because both $k$NN and TabPFN yield a conformal threshold of $\hat{q}\approx 1.0$, this single maximum-probability class immediately satisfies the APS stopping rule for $k$NN. This keeps its mean set size near one, even though the probability of 1.0 is a discrete artifact of neighbor voting rather than a calibrated confidence measure. Furthermore, when $k$NN is confidently wrong, this same mechanism produces size-1 sets that fail to capture the true label, resulting in lower empirical conformal coverage ($94.9\%$ vs.\ $99.7\%$ for TabPFN).

Crucially, the $90\%$ conformal target is a mathematical lower-bound guarantee, not a strict upper limit; TabPFN's $99.7\%$ empirical coverage indicates conservative over-coverage rather than a calibration failure. TabPFN distributes probability mass across the non-top-1 classes, typically yielding a high maximum predicted probability that remains below 1.0. To reach the conformal threshold $\hat{q} \approx 1.0$, APS for TabPFN must therefore accumulate probabilities from several classes, resulting in a larger average prediction set size (mean ASS $\approx 5.5$). So, this deterministic APS adds classes from the most to the least probable until their summed probability reaches $\hat{q}$ (or 90\%). It always keeps the last class added, even when only a small part of that class's probability was needed to reach $\hat{q}$, which enlarges the ASS. In Appendix~G (Table~\ref{tab:appendix_aps_randomized}), we show that randomized APS~\citep{romano2020classification} aligns all heads to the $90\%$ target coverage ($91.4\%$ for TabPFN) and reduces TabPFN's mean set size from $5.51$ to $1.39$, tying with logistic regression ($1.39$) for the smallest prediction sets at $k=100$.

\begin{table}[t]
\centering
\caption{Calibration versus conformal prediction-set efficiency at the canonical setting (mean $\pm$ SEM over 560 ablation episodes). Top-$k$ ECE bins cumulative top-$k$ confidence; ASS is the mean set size from split conformal APS ($\alpha=0.1$). Bold = best per column.}
\label{tab:ablation_ass}
\begin{adjustbox}{max width=\columnwidth}
\footnotesize
\begin{tabular}{lcccc}
\hline
\textbf{Head} & \textbf{Top-1 ECE} & \textbf{Top-5 ECE} & \textbf{ASS} & \textbf{Conf.\ cov.} \\
\hline
TabPFN        & \textbf{0.031 $\pm$ 0.003} & \textbf{0.005 $\pm$ 0.001} & 5.46 $\pm$ 0.06 & \textbf{99.7 $\pm$ 0.1}\% \\
$k$NN         & 0.037 $\pm$ 0.003 & 0.038 $\pm$ 0.004 & \textbf{1.80 $\pm$ 0.08} & 94.9 $\pm$ 0.6\% \\
Logistic Reg. & 0.069 $\pm$ 0.008 & 0.013 $\pm$ 0.002 & 5.96 $\pm$ 0.06 & 99.5 $\pm$ 0.1\% \\
Linear SVM    & 0.089 $\pm$ 0.003 & 0.029 $\pm$ 0.001 & 5.32 $\pm$ 0.05 & 99.5 $\pm$ 0.1\% \\
\hline
\end{tabular}
\end{adjustbox}
\end{table}

\subsection{Ablation 3: ICL Models and Calibration}
\label{sec:ablation_pfn_heads}

Experiments~1--3 establish that TabPFN~v2.5~\citep{grinsztajn2025tabpfn} yields well-calibrated posteriors on frozen multimodal embeddings without gradient updates. Because TabPFN~v2.6 and v3~\citep{grinsztajn2026tabpfn} were released after our main experiments concluded, we evaluate them here alongside other ICL models to determine whether the calibration advantage is unique to TabPFN or a shared property of this broader model class. We re-run the canonical setting ($C=10$, $k=100$, $d=96$) with seven ICL foundation models: TabPFN~v2~\citep{hollmann2025accurate}, v2.5, v2.6, and v3~\citep{grinsztajn2026tabpfn}, TabICL~v1 and v2~\citep{qu2025tabicl,qu2026tabiclv2}, and TabDPT~\citep{ma2025tabdpt}. These are compared against the three primary classical heads from Experiment~1 ($k$NN, logistic regression, linear SVM). All ICL models process the same $(X_{\mathrm{train}}, y_{\mathrm{train}})$ context in a single forward pass at inference time, whereas the classical heads are fit to that context using their standard algorithms.

Table~\ref{tab:ablation5_pfn} shows that, collectively, the evaluated ICL foundation models demonstrate strong calibration performance. Averaged over $1{,}400$ episodes, all ICL models except the earliest version (TabPFN~v2) achieve a lower ECE than the best classical head ($k$NN at $0.037 \pm 0.003$), reaching as low as $0.019 \pm 0.001$ for TabDPT and TabICL~v2. Furthermore, every ICL model achieves strictly lower NLL than all classical methods; the NLL for ICL models remains at or below $0.321$, whereas classical methods range from $0.399$ (linear SVM) to $1.727$ ($k$NN). The Top-$k$ curves in Appendix Figure~\ref{fig:ablation5_topkp} extend this pattern to $k=5$, where ICL models consistently yield higher Top-$k$ accuracy and maintain lower Top-$k$ ECE across almost every rank. Overall, these results suggest that transferring calibration to frozen encoder features is a shared strength of the evaluated ICL foundation models, rather than a phenomenon unique to TabPFN~v2.5.x.

Table~\ref{tab:ablation5_pfn} also highlights the trade-off between checkpoint size and marginal performance gains. While TabDPT, TabICL~v2, and TabPFN~v3 lead slightly overall, these improvements require checkpoints $3$ to $6\times$ larger than TabPFN~v2.5 ($43$\,MB) for less than a $1$ percentage-point accuracy increase. Because v2.5 effectively balances this memory trade-off while already sitting on the calibration frontier, we retain it as the default head for our primary experiments.

\begin{table}[t]
\centering
\caption{ICL foundation-model heads at the canonical setting (mean $\pm$ SEM over $1{,}400$ episodes). Size: released classifier checkpoint. Bold = best per column among all heads.}
\label{tab:ablation5_pfn}
\begin{adjustbox}{max width=\columnwidth}
\footnotesize
\begin{tabular}{lccccc}
\hline
\textbf{Head} & \textbf{Size (MB)} & \textbf{Acc} & \textbf{ECE} & \textbf{NLL} & \textbf{ASS} \\
\hline
TabPFN v2     & 29  & 0.898 $\pm$ 0.010 & 0.045 $\pm$ 0.005 & 0.321 $\pm$ 0.030 & 5.11 $\pm$ 0.06 \\
TabPFN v2.5   & 43  & 0.905 $\pm$ 0.010 & 0.031 $\pm$ 0.003 & 0.287 $\pm$ 0.028 & 5.45 $\pm$ 0.06 \\
TabPFN v2.6   & 43  & 0.896 $\pm$ 0.010 & 0.036 $\pm$ 0.003 & 0.316 $\pm$ 0.030 & 5.29 $\pm$ 0.05 \\
TabPFN v3     & 213 & 0.901 $\pm$ 0.010 & 0.021 $\pm$ 0.002 & 0.286 $\pm$ 0.028 & 3.62 $\pm$ 0.12 \\
TabICL v1     & 108 & 0.910 $\pm$ 0.009 & 0.022 $\pm$ 0.002 & 0.272 $\pm$ 0.027 & 5.48 $\pm$ 0.05 \\
TabICL v2     & 110 & 0.913 $\pm$ 0.009 & \textbf{0.019} $\pm$ 0.001 & 0.254 $\pm$ 0.026 & 5.37 $\pm$ 0.06 \\
TabDPT        & 254 & \textbf{0.914 $\pm$ 0.009} & \textbf{0.019 $\pm$ 0.001} & \textbf{0.253 $\pm$ 0.026} & 6.06 $\pm$ 0.04 \\
$k$NN     & --- & 0.848 $\pm$ 0.015 & 0.037 $\pm$ 0.003 & 1.727 $\pm$ 0.184 & \textbf{1.80 $\pm$ 0.08} \\
Logistic Reg.     & --- & 0.889 $\pm$ 0.011 & 0.069 $\pm$ 0.008 & 0.581 $\pm$ 0.068 & 5.97 $\pm$ 0.06 \\
Linear SVM   & --- & 0.885 $\pm$ 0.011 & 0.088 $\pm$ 0.002 & 0.399 $\pm$ 0.032 & 5.32 $\pm$ 0.05 \\
\hline
\end{tabular}
\end{adjustbox}
\end{table}

\section{Discussion}
\label{sec:discussion}

\paragraph{Interpretation.} Although images, text, and audio are not tabular inputs, their pooled encoder outputs are fixed-length numerical vectors. The encoder has already performed the modality-specific representation learning, so TabPFN only needs to infer a decision rule from these numerical features. Standardization and PCA reduce redundancy and constrain their dimension, potentially making the resulting classification problem closer to the moderate-dimensional structures represented during TabPFN pretraining. Its posterior-predictive objective may then regularize confidence more effectively than heads trained primarily for discrimination. This explanation is consistent with our results but is not a causally verified mechanism.

The operating boundary supports this interpretation. TabPFN benefits from enough labeled context relative to the representation dimension, while extremely low shot counts and high dimensions weaken its accuracy advantage. Multiclass tasks may provide more structure for in-context inference, whereas near-ceiling tasks leave little room for improvement. Backbone adaptation and head choice are also complementary: adaptation improves task-specific representations, whereas TabPFN improves the reliability of predictions made from those representations. The targeted adaptation study does not establish that this pattern holds across all backbones and domains. However, lower NLL and ECE do not guarantee compact conformal sets. TabPFN's larger ASS reflects this separate trade-off. Its smoother probability distributions require longer prefixes to cross the APS threshold, producing larger sets and conservative over-coverage.

\paragraph{Practical guidance.} TabPFN is a strong frozen-encoder head when calibrated probabilities matter and embeddings can be reduced to a moderate dimension. Logistic regression or SVM remains appropriate for extremely low shot counts, high dimensions, or near-ceiling tasks. For a mismatched encoder, the results support adapting the backbone and then replacing its trained head with TabPFN. Applications requiring the smallest conformal sets should account for TabPFN's larger ASS.

\paragraph{Limitations.} The proposed explanations are interpretive rather than mechanistically tested. Backbone adaptation uses a restricted encoder subset and fixed training schedule, and PCA becomes costly at large sample sizes and dimensions. Our evaluation is limited to in-distribution discrete classification, so the findings may not generalize to domain shift, regression, retrieval, or structured prediction.

\section{Conclusion}
\label{sec:conclusion}

Can TabPFN provide reliable confidence on multimodal embeddings without sacrificing predictive accuracy? Our results give a conditional answer. Across image, text, and audio encoders, TabPFN broadly reduces NLL and ECE while retaining competitive accuracy. Calibration and accuracy gains occur together most consistently at moderate-to-high shot counts and low-to-moderate dimensions, with less accuracy benefit when labels are scarce, dimensions are high, or baselines approach ceiling performance. After backbone adaptation, replacing the trained head with TabPFN improves probability reliability while preserving or improving the adapted model's accuracy in most configurations. These findings provide a unified account of when TabPFN is useful for calibration-sensitive multimodal classification.

\clearpage  

\bibliography{references}

\clearpage  
\appendix
\onecolumn

\section{Appendix}
\label{sec:appendix}

Appendix~A shows per-head reliability diagrams that visualize the shape of miscalibration at the mid-shot setting $k=100$. Appendix~B provides NLL and ASS as additional metrics complementing the Accuracy and ECE metrics in Figure~\ref{fig:exp2}. Appendix~C shows the full $\Delta$ACC distribution for ceiling-effect analysis. Appendix~D reports stratified performance means for all nine heads at $C=10$. Appendix~E reports the full two-dimensional $(k\times d)$ win-rate map behind the curves of Figure~\ref{fig:exp2}, along with controlled fixed-task-subset evaluations at $C=10$. Appendix~F provides NLL and ASS summaries, temperature-scaled evaluations under an equal $80\%$ label split, and AudioMNIST reliability diagrams for the fine-tuning experiment. Appendix~G provides extended Top-$k$ calibration diagnostics and randomized conformal prediction set (APS) evaluations for Ablation~2. Appendix~H presents the Top-$k$ calibration and accuracy comparison across the broader class of ICL foundation models evaluated in Ablation~3. Appendix~I compares TabPFN against post-hoc temperature-scaled baselines across shot counts. Appendix~J compares PCA against alternative dimensionality reduction methods and evaluates the feature-compression trade-off at $d=96$.

\subsection{Appendix A: Reliability Diagrams}
\label{sec:appendix_a}

\begin{figure}[htbp]
\centering
\includegraphics[width=0.6\columnwidth]{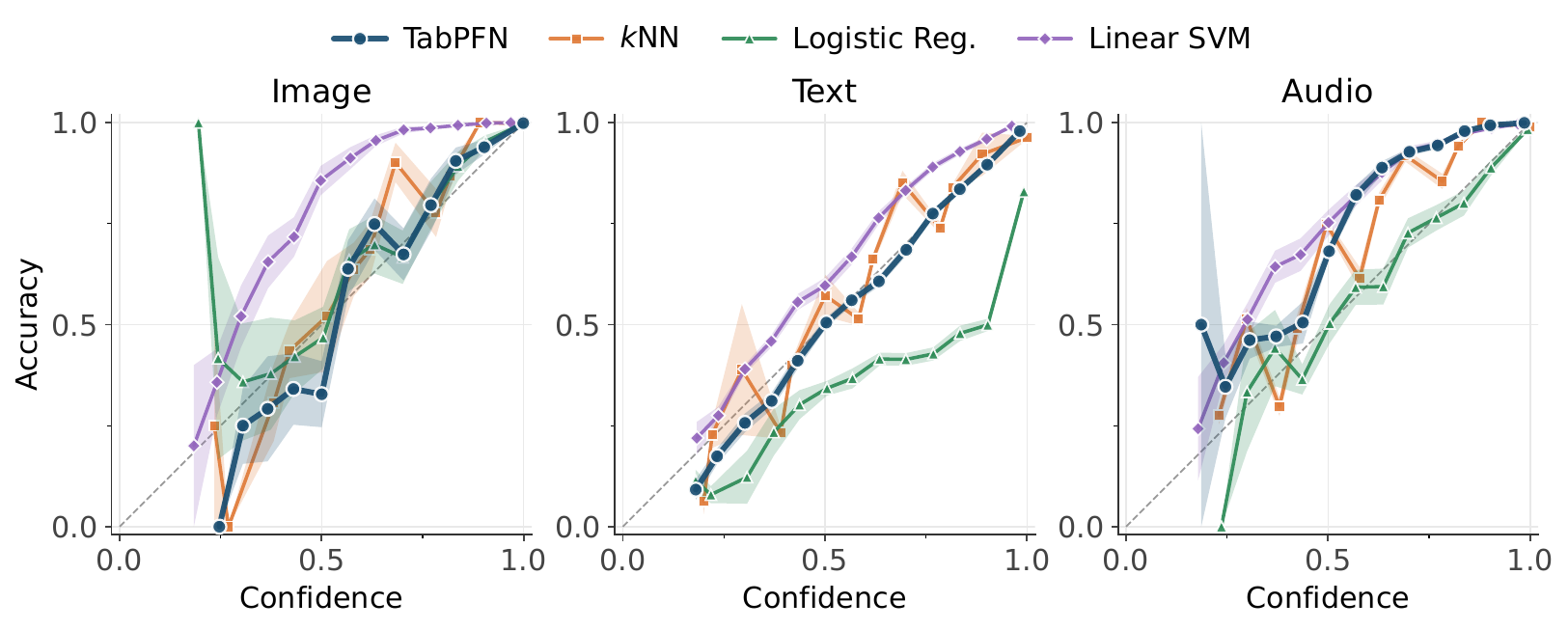}
\caption{%
Reliability diagrams at $k=100$, $d=96$, $C=10$ (mean $\pm$1 SEM across 5 seeds). Dashed diagonal: perfect calibration.%
}
\label{fig:appendix1_reliability}
\end{figure}

Figure~\ref{fig:appendix1_reliability} shows reliability diagrams for the four primary heads at the canonical setting ($k=100$, $d=96$, $C=10$), one panel per modality. Each curve plots mean bin accuracy against mean bin confidence across 5 seeds; shaded bands indicate $\pm$1 SEM. The diagonal $y=x$ is the perfect-calibration reference. Curves below the diagonal indicate over-confidence (stated confidence exceeds empirical accuracy); curves above indicate under-confidence. TabPFN tracks the diagonal better than other classification heads, and the gap is most pronounced in the Text panel (20 Newsgroups), consistent with TabPFN's lowest ECE rank in Table~\ref{tab:exp1_rank}.

\subsection{Appendix B: NLL and ASS Results}
\label{sec:appendix_b}

\begin{figure}[htbp]
\centering
\includegraphics[width=0.55\columnwidth]{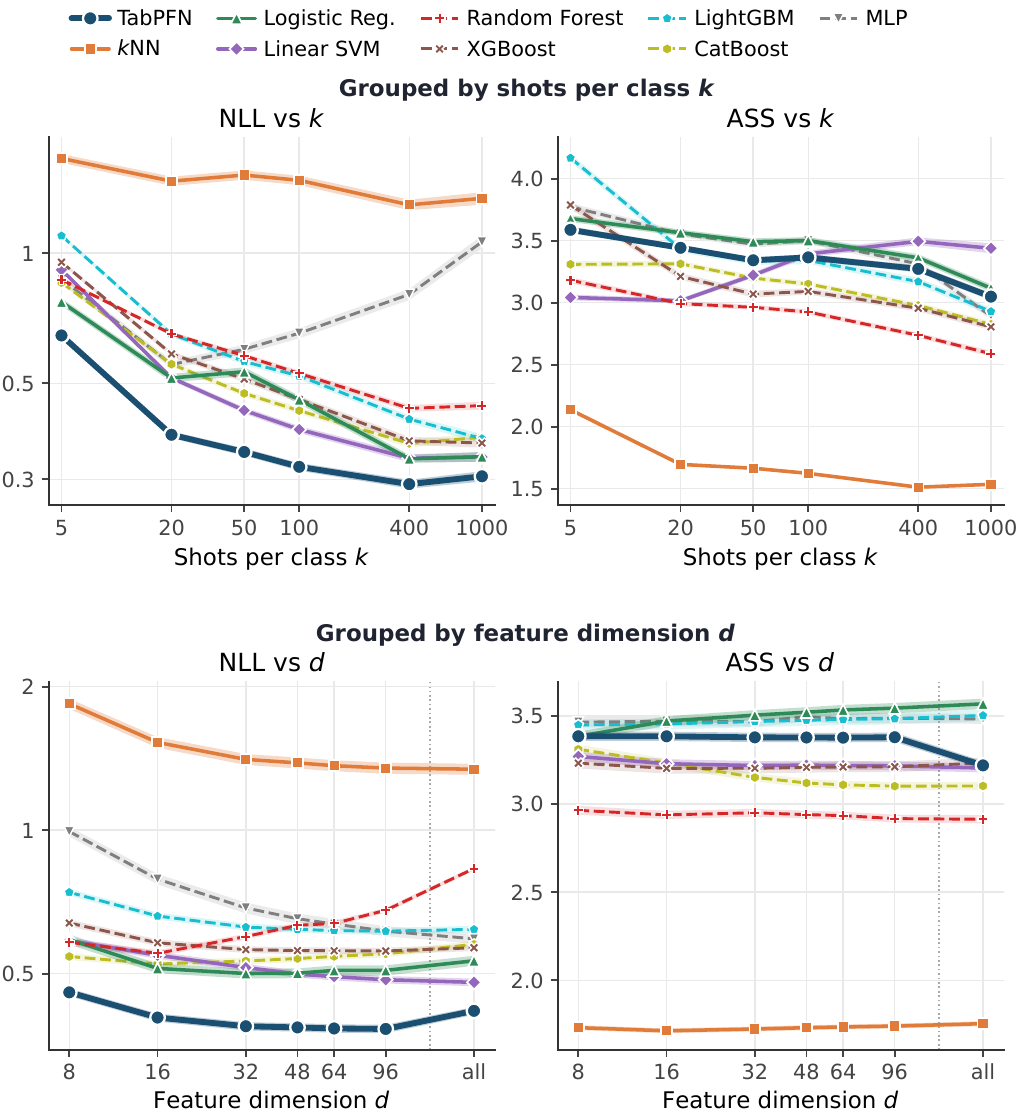}
\caption{%
NLL and ASS for all nine heads versus shots per class $k$ (top) and feature dimension $d$ (bottom), aggregated over all datasets, encoders, seeds, and $C\in\{2,5,10\}$. Shaded regions indicate $\pm$1 SEM.%
}
\label{fig:exp2_appendix_nll_ass}
\end{figure}

Figure~\ref{fig:exp2_appendix_nll_ass} reports NLL and ASS, complementing the Accuracy and ECE panels shown in Figure~\ref{fig:exp2}. On NLL (log scale), TabPFN achieves strictly lower values than all baselines at every $k$ and $d$, mirroring the ECE pattern. ASS remains the only metric on which TabPFN does not rank first.

\subsection{Appendix C: Accuracy Gains and the Ceiling Effect}
\label{sec:appendix_c}

\begin{figure}[htbp]
\centering
\includegraphics[width=0.5\columnwidth]{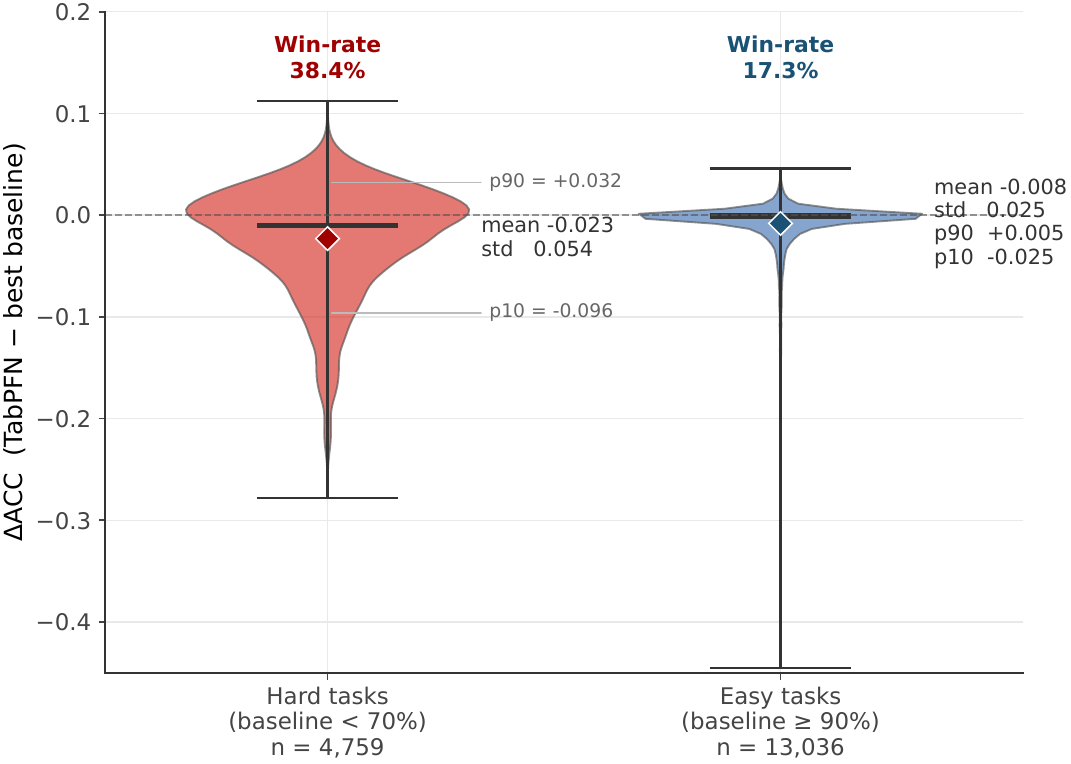}
\caption{%
Distribution of $\Delta$ACC (TabPFN $-$ best-of-8) for hard (baseline $<70\%$, $n=4{,}759$) versus easy ($\geq 90\%$, $n=13{,}036$) tasks.%
}
\label{fig:exp2_appendix_difficulty}
\end{figure}

Figure~\ref{fig:exp2_appendix_difficulty} shows the full $\Delta$ACC distribution for hard and easy tasks. For easy tasks, the distribution is sharply concentrated near zero (std $0.025$), whereas for hard tasks the distribution is approximately twice as wide (std $0.054$) with a more pronounced tail. The higher win-rate on hard tasks (38.4\% vs.\ 17.3\% on easy tasks) reflects the ceiling effect: when the best baseline already approaches perfect accuracy, the margin available for any competing method to improve is negligible, suppressing win-rates regardless of method quality.

\subsection{Appendix D: Stratified Evaluation Metrics}
\label{sec:appendix_d}

Tables~\ref{tab:exp2_shots_stratified} and~\ref{tab:exp2_pca_stratified} report the mean $\pm$ standard deviation for all nine classification heads at $C=10$, complementing the marginal curves presented in Figure~\ref{fig:exp2}. We partition the evaluation grid into three sample-size regimes: Low-$k$ ($\{5, 20\}$), Mid-$k$ ($\{50, 100\}$), and High-$k$ ($\{400, 1000\}$). Similarly, the feature dimensions are grouped into three regimes: Low-$d$ ($\{8, 16\}$), Mid-$d$ ($\{32, 48\}$), and High-$d$ ($\{64, 96\}$). Values in both tables are aggregated over all datasets, encoders, and seeds, excluding the $d=\texttt{all}$ configuration to prevent its capacity-limit degradation from artificially influencing the averages. Bold text indicates the best mean performance per column among all nine heads. Overall, TabPFN achieves the lowest NLL and ECE in every regime; on accuracy, it outperforms logistic regression at Mid-$k$ and High-$k$, though it trails at Low-$k$, Mid-$d$, and High-$d$, while $k$NN yields the lowest ASS value.

\begin{table*}[t]
\centering
\caption{Acc / NLL / ECE / ASS stratified by number of shots per class $k$ at $C=10$ (mean $\pm$ std over all datasets, encoders, and PCA dims). Buckets: Low-$k$ $k\in\{5,20\}$; Mid-$k$ $k\in\{50,100\}$; High-$k$ $k\in\{400,1000\}$. Lower is better for NLL, ECE, and ASS; higher is better for Acc. Bold = best per column.}
\label{tab:exp2_shots_stratified}
\begin{adjustbox}{max width=\textwidth}
\scriptsize
\setlength{\tabcolsep}{3pt}
\begin{tabular}{l|ccc|ccc|ccc|ccc}
\hline
& \multicolumn{3}{c|}{\textbf{Acc}} & \multicolumn{3}{c|}{\textbf{NLL}} & \multicolumn{3}{c|}{\textbf{ECE}} & \multicolumn{3}{c|}{\textbf{ASS}} \\
\textbf{Head} & Low-$k$ & Mid-$k$ & High-$k$ & Low-$k$ & Mid-$k$ & High-$k$ & Low-$k$ & Mid-$k$ & High-$k$ & Low-$k$ & Mid-$k$ & High-$k$ \\
\hline
TabPFN          & 0.783 $\pm$ 0.224 & \textbf{0.868 $\pm$ 0.157} & \textbf{0.923 $\pm$ 0.099} & \textbf{0.640 $\pm$ 0.643} & \textbf{0.385 $\pm$ 0.440} & \textbf{0.224 $\pm$ 0.271} & \textbf{0.047 $\pm$ 0.042} & \textbf{0.027 $\pm$ 0.024} & \textbf{0.019 $\pm$ 0.016} & 5.629 $\pm$ 1.011 & 5.405 $\pm$ 0.685 & 5.295 $\pm$ 0.728 \\
$k$NN           & 0.784 $\pm$ 0.232 & 0.819 $\pm$ 0.204 & 0.897 $\pm$ 0.115 & 2.550 $\pm$ 3.201 & 2.170 $\pm$ 2.712 & 1.131 $\pm$ 1.329 & 0.063 $\pm$ 0.050 & 0.043 $\pm$ 0.043 & 0.033 $\pm$ 0.030 & \textbf{2.250 $\pm$ 1.150} & \textbf{1.866 $\pm$ 0.968} & \textbf{1.469 $\pm$ 0.509} \\
Logistic Reg.   & \textbf{0.812 $\pm$ 0.210} & 0.853 $\pm$ 0.170 & 0.903 $\pm$ 0.118 & 0.766 $\pm$ 0.964 & 0.564 $\pm$ 0.676 & 0.294 $\pm$ 0.327 & 0.089 $\pm$ 0.097 & 0.055 $\pm$ 0.070 & 0.022 $\pm$ 0.017 & 5.864 $\pm$ 1.032 & 5.735 $\pm$ 0.899 & 5.391 $\pm$ 0.894 \\
Linear SVM      & 0.798 $\pm$ 0.216 & 0.845 $\pm$ 0.174 & 0.901 $\pm$ 0.119 & 1.019 $\pm$ 0.520 & 0.533 $\pm$ 0.481 & 0.297 $\pm$ 0.328 & 0.325 $\pm$ 0.145 & 0.115 $\pm$ 0.041 & 0.036 $\pm$ 0.025 & 4.250 $\pm$ 1.548 & 5.276 $\pm$ 0.755 & 6.081 $\pm$ 0.870 \\
Random Forest   & 0.766 $\pm$ 0.229 & 0.830 $\pm$ 0.185 & 0.894 $\pm$ 0.114 & 1.047 $\pm$ 0.635 & 0.729 $\pm$ 0.597 & 0.471 $\pm$ 0.423 & 0.274 $\pm$ 0.134 & 0.189 $\pm$ 0.117 & 0.142 $\pm$ 0.136 & 4.477 $\pm$ 1.619 & 4.292 $\pm$ 1.114 & 3.652 $\pm$ 1.104 \\
XGBoost         & 0.685 $\pm$ 0.226 & 0.815 $\pm$ 0.189 & 0.895 $\pm$ 0.113 & 1.030 $\pm$ 0.667 & 0.579 $\pm$ 0.560 & 0.318 $\pm$ 0.318 & 0.107 $\pm$ 0.066 & 0.049 $\pm$ 0.047 & 0.031 $\pm$ 0.027 & 5.916 $\pm$ 1.510 & 5.042 $\pm$ 1.321 & 4.473 $\pm$ 1.179 \\
LightGBM        & 0.732 $\pm$ 0.231 & 0.828 $\pm$ 0.182 & 0.902 $\pm$ 0.110 & 0.958 $\pm$ 0.830 & 0.680 $\pm$ 0.709 & 0.350 $\pm$ 0.353 & 0.105 $\pm$ 0.098 & 0.092 $\pm$ 0.096 & 0.044 $\pm$ 0.045 & 6.216 $\pm$ 1.167 & 5.733 $\pm$ 1.302 & 5.086 $\pm$ 1.265 \\
CatBoost        & 0.772 $\pm$ 0.228 & 0.833 $\pm$ 0.179 & 0.892 $\pm$ 0.115 & 1.008 $\pm$ 0.660 & 0.594 $\pm$ 0.552 & 0.362 $\pm$ 0.343 & 0.266 $\pm$ 0.142 & 0.116 $\pm$ 0.090 & 0.068 $\pm$ 0.076 & 5.114 $\pm$ 1.309 & 4.987 $\pm$ 0.734 & 4.495 $\pm$ 0.619 \\
MLP             & 0.780 $\pm$ 0.224 & 0.848 $\pm$ 0.175 & 0.906 $\pm$ 0.119 & 0.882 $\pm$ 0.994 & 0.801 $\pm$ 1.248 & 0.543 $\pm$ 0.721 & 0.104 $\pm$ 0.105 & 0.094 $\pm$ 0.123 & 0.064 $\pm$ 0.078 & 5.934 $\pm$ 1.115 & 5.680 $\pm$ 0.809 & 5.072 $\pm$ 1.223 \\
\hline
\end{tabular}
\end{adjustbox}
\end{table*}

\begin{table*}[t]
\centering
\caption{Acc / NLL / ECE / ASS stratified by feature dimension $d$ at $C=10$ (mean $\pm$ std over all datasets, encoders, and shots per class $k$). Buckets: Low-$d$ $d\in\{8,16\}$; Mid-$d$ $d\in\{32,48\}$; High-$d$ $d\in\{64,96\}$; $d=\texttt{all}$ excluded. Lower is better for NLL, ECE, and ASS; higher is better for Acc. Bold = best per column.}
\label{tab:exp2_pca_stratified}
\begin{adjustbox}{max width=\textwidth}
\scriptsize
\setlength{\tabcolsep}{3pt}
\begin{tabular}{l|ccc|ccc|ccc|ccc}
\hline
& \multicolumn{3}{c|}{\textbf{Acc}} & \multicolumn{3}{c|}{\textbf{NLL}} & \multicolumn{3}{c|}{\textbf{ECE}} & \multicolumn{3}{c|}{\textbf{ASS}} \\
\textbf{Head} & Low-$d$ & Mid-$d$ & High-$d$ & Low-$d$ & Mid-$d$ & High-$d$ & Low-$d$ & Mid-$d$ & High-$d$ & Low-$d$ & Mid-$d$ & High-$d$ \\
\hline
TabPFN          & \textbf{0.810 $\pm$ 0.206} & 0.854 $\pm$ 0.181 & 0.858 $\pm$ 0.181 & \textbf{0.541 $\pm$ 0.574} & \textbf{0.434 $\pm$ 0.524} & \textbf{0.426 $\pm$ 0.525} & \textbf{0.031 $\pm$ 0.026} & \textbf{0.035 $\pm$ 0.036} & \textbf{0.038 $\pm$ 0.040} & 5.469 $\pm$ 0.907 & 5.494 $\pm$ 0.833 & 5.489 $\pm$ 0.855 \\
$k$NN           & 0.789 $\pm$ 0.227 & 0.832 $\pm$ 0.198 & 0.836 $\pm$ 0.195 & 2.604 $\pm$ 3.193 & 1.936 $\pm$ 2.580 & 1.861 $\pm$ 2.503 & 0.055 $\pm$ 0.054 & 0.047 $\pm$ 0.041 & 0.048 $\pm$ 0.042 & \textbf{1.980 $\pm$ 1.050} & \textbf{1.943 $\pm$ 1.022} & \textbf{1.955 $\pm$ 1.022} \\
Logistic Reg.   & 0.800 $\pm$ 0.214 & \textbf{0.861 $\pm$ 0.167} & \textbf{0.872 $\pm$ 0.159} & 0.731 $\pm$ 0.934 & 0.554 $\pm$ 0.729 & 0.517 $\pm$ 0.679 & 0.063 $\pm$ 0.090 & 0.065 $\pm$ 0.079 & 0.064 $\pm$ 0.075 & 5.464 $\pm$ 1.100 & 5.811 $\pm$ 0.907 & 5.897 $\pm$ 0.849 \\
Linear SVM      & 0.791 $\pm$ 0.217 & 0.851 $\pm$ 0.174 & 0.863 $\pm$ 0.166 & 0.817 $\pm$ 0.610 & 0.669 $\pm$ 0.532 & 0.628 $\pm$ 0.512 & 0.191 $\pm$ 0.165 & 0.199 $\pm$ 0.158 & 0.192 $\pm$ 0.151 & 5.016 $\pm$ 1.484 & 4.953 $\pm$ 1.340 & 4.956 $\pm$ 1.351 \\
Random Forest   & 0.785 $\pm$ 0.220 & 0.828 $\pm$ 0.190 & 0.829 $\pm$ 0.191 & 0.778 $\pm$ 0.668 & 0.811 $\pm$ 0.604 & 0.873 $\pm$ 0.602 & 0.128 $\pm$ 0.090 & 0.243 $\pm$ 0.136 & 0.283 $\pm$ 0.137 & 4.295 $\pm$ 1.563 & 4.245 $\pm$ 1.292 & 4.202 $\pm$ 1.320 \\
XGBoost         & 0.747 $\pm$ 0.225 & 0.784 $\pm$ 0.204 & 0.786 $\pm$ 0.205 & 0.804 $\pm$ 0.691 & 0.695 $\pm$ 0.615 & 0.691 $\pm$ 0.612 & 0.083 $\pm$ 0.074 & 0.066 $\pm$ 0.056 & 0.065 $\pm$ 0.055 & 5.321 $\pm$ 1.513 & 5.312 $\pm$ 1.490 & 5.332 $\pm$ 1.492 \\
LightGBM        & 0.767 $\pm$ 0.225 & 0.815 $\pm$ 0.195 & 0.817 $\pm$ 0.193 & 0.889 $\pm$ 0.852 & 0.675 $\pm$ 0.693 & 0.653 $\pm$ 0.671 & 0.112 $\pm$ 0.110 & 0.079 $\pm$ 0.082 & 0.074 $\pm$ 0.075 & 5.737 $\pm$ 1.285 & 5.848 $\pm$ 1.302 & 5.878 $\pm$ 1.328 \\
CatBoost        & 0.792 $\pm$ 0.217 & 0.831 $\pm$ 0.187 & 0.830 $\pm$ 0.187 & 0.723 $\pm$ 0.622 & 0.726 $\pm$ 0.624 & 0.753 $\pm$ 0.638 & 0.123 $\pm$ 0.112 & 0.192 $\pm$ 0.147 & 0.207 $\pm$ 0.150 & 5.241 $\pm$ 1.007 & 4.846 $\pm$ 1.003 & 4.750 $\pm$ 1.056 \\
MLP             & 0.795 $\pm$ 0.219 & 0.844 $\pm$ 0.183 & 0.848 $\pm$ 0.180 & 1.103 $\pm$ 1.440 & 0.652 $\pm$ 0.774 & 0.603 $\pm$ 0.710 & 0.126 $\pm$ 0.144 & 0.079 $\pm$ 0.085 & 0.072 $\pm$ 0.075 & 5.672 $\pm$ 1.330 & 5.672 $\pm$ 0.963 & 5.674 $\pm$ 0.936 \\
\hline
\end{tabular}
\end{adjustbox}
\end{table*}

\subsection{Appendix E: Full $(k\times d)$ Win-Rate Map}
\label{sec:appendix_e}

\begin{figure}[htbp]
\centering
\includegraphics[width=1.0\textwidth]{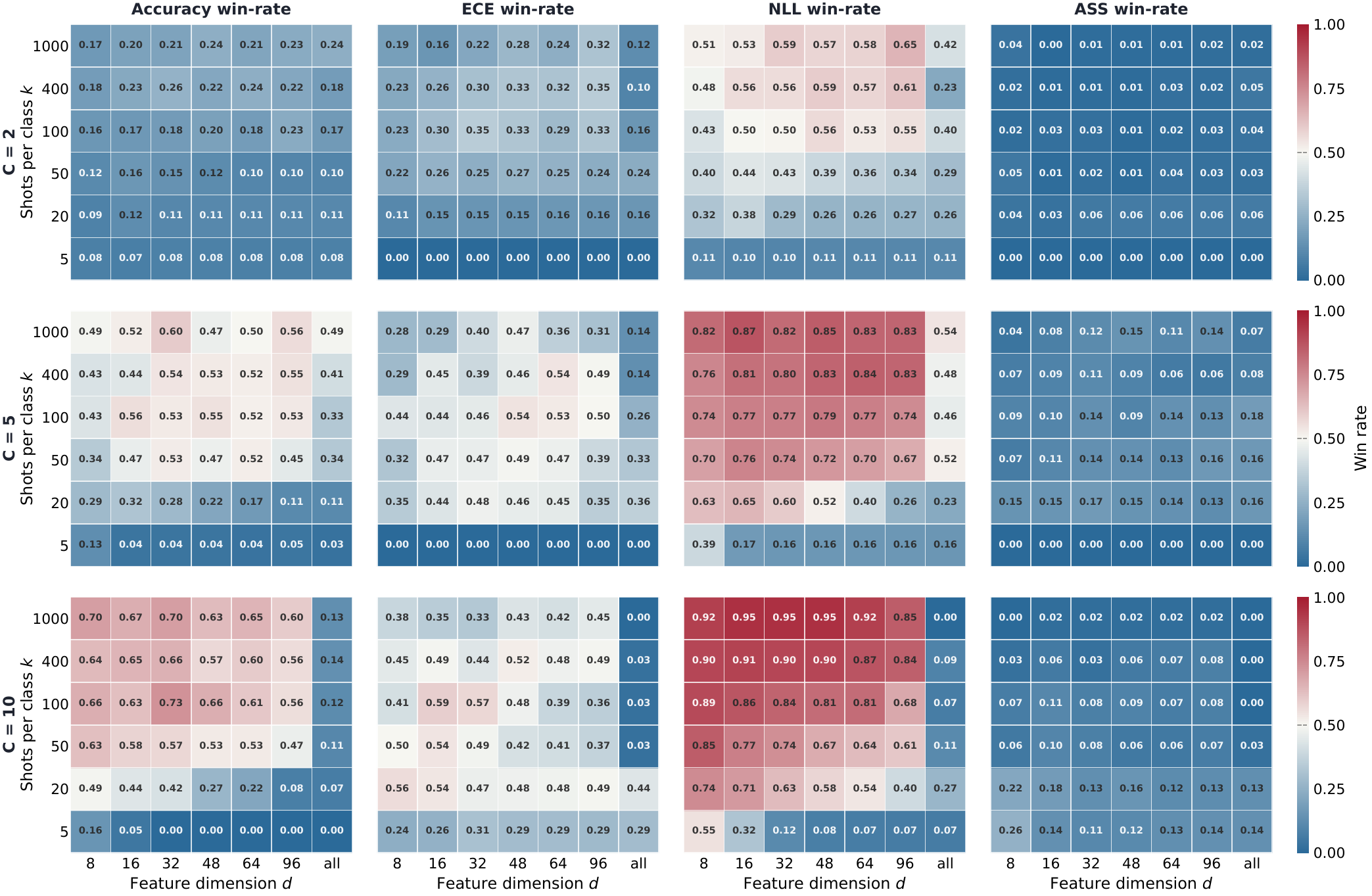}
\caption{%
Full $(k\times d)$ win-rate map. Rows: $C\in\{2,5,10\}$; columns: Accuracy, ECE, NLL, ASS. Each episode is $P(\text{TabPFN}>\text{best-of-8})$ at that $(k,d)$.
}
\label{fig:exp2_appendix_heatmaps}
\end{figure}

Figure~\ref{fig:exp2_appendix_heatmaps} presents the accuracy--reliability boundary in two dimensions simultaneously. In the Accuracy column, the TabPFN-wins regime is nearly absent at $C=2$, emerges as a mid-$k$ band at $C=5$, and broadens into an extended high-win-rate regime at $C=10$, consistent with the accuracy improvement with class count in Figure~\ref{fig:exp2}. In every row, the $d=\texttt{all}$ column shows near-zero win rates, indicating that TabPFN's relative accuracy advantage does not extend to the maximum available PCA dimension. In the NLL column, TabPFN achieves an overall mean win-rate of $0.53$; however, while win-rates are predominantly above $0.5$ for multiclass tasks ($C \in \{5, 10\}$), they fall slightly below this threshold for binary tasks ($C=2$). The ASS column shows uniformly low win-rates (mean $0.07$).

\paragraph{Controlled fixed-task-subset grid at $C=10$.}
In the full evaluation grid (Figure~\ref{fig:exp2} and Figure~\ref{fig:exp2_appendix_heatmaps}), smaller datasets have a limited number of examples per class and cannot support large shot counts; consequently, several datasets included at $k=100$ drop out at $k=400$ and $1000$. In addition, when $k$ is very small, the training sample size caps the maximum PCA dimension at $d_{\mathrm{used}} = \min(d,\, kC - 1,\, d_{\mathrm{enc}},\, 2000)$, causing multiple large $d$ settings to reduce to the same small value. Because both factors could shift the aggregated curves in Figure~\ref{fig:exp2}, we re-evaluate the $C=10$ grid across all nine uncalibrated heads (5 seeds) using two fixed groups of dataset--encoder pairs: $\mathcal{S}_{26}$, comprising the 26 pairs that support every $k \le 100$, and $\mathcal{S}_{11}$, comprising the 11 pairs with larger dataset size that support every $k$ up to $400$ and $1000$. Within each group, we also count reduced PCA dimensions only once. For example, at $k=5$ and $C=10$ ($kC - 1 = 49$), dimension settings $d \in \{64, 96, \texttt{all}\}$ all reduce to $49$ features and are counted together in $d=\texttt{all}$ cell.

\begin{figure}[htbp]
\centering
\includegraphics[width=0.98\textwidth]{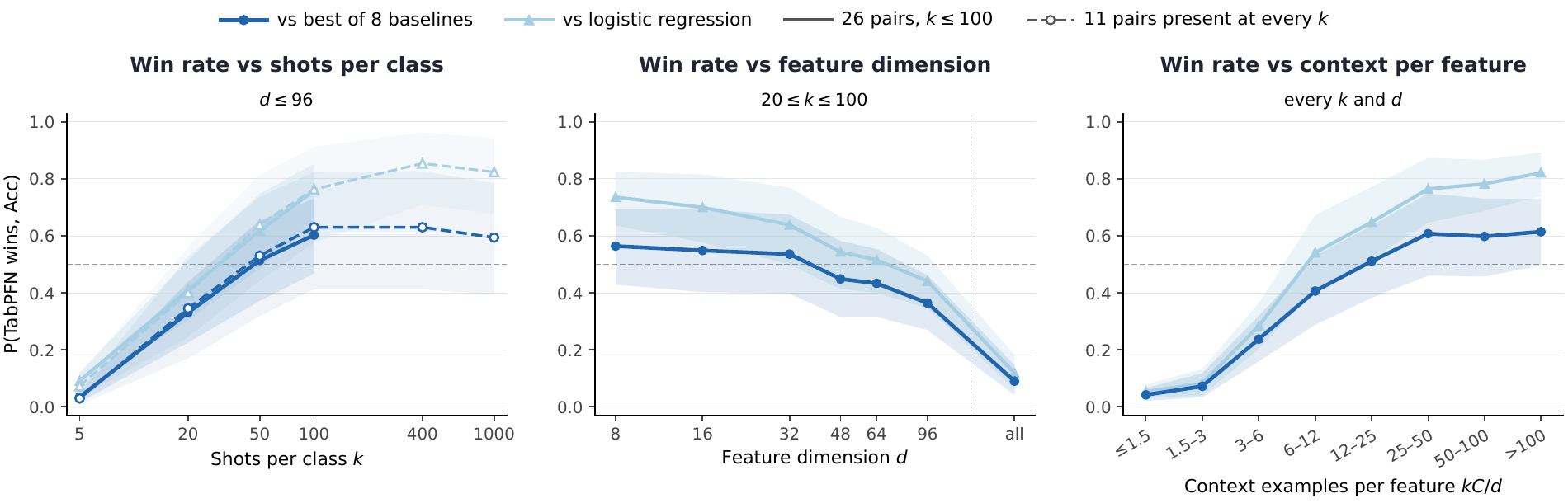}
\caption{%
Fixed-subset accuracy win rate of TabPFN at $C=10$, against the best of the eight baselines in each episode and against logistic regression (LR) alone, with shaded area indicating 95\% bootstrap confidence intervals over unique dataset--encoder pairs. \emph{Left:} win rate versus shots per class $k$ ($d \le 96$) on the 26-pair subset $\mathcal{S}_{26}$ ($k \le 100$) and the 11-pair subset $\mathcal{S}_{11}$ ($k \le 1000$, dashed). \emph{Middle:} win rate versus feature dimension $d$ ($20 \le k \le 100$ on $\mathcal{S}_{26}$). \emph{Right:} win rate versus context examples per used feature, $kC/d$, pooling the $\mathcal{S}_{26}$ episodes at $k \le 100$ and the $\mathcal{S}_{11}$ episodes at $k \in \{400, 1000\}$ into one curve (all $d$ are used).%
}
\label{fig:appendix_fixed_grid_winrate}
\end{figure}

\begin{figure}[htbp]
\centering
\includegraphics[width=0.98\textwidth]{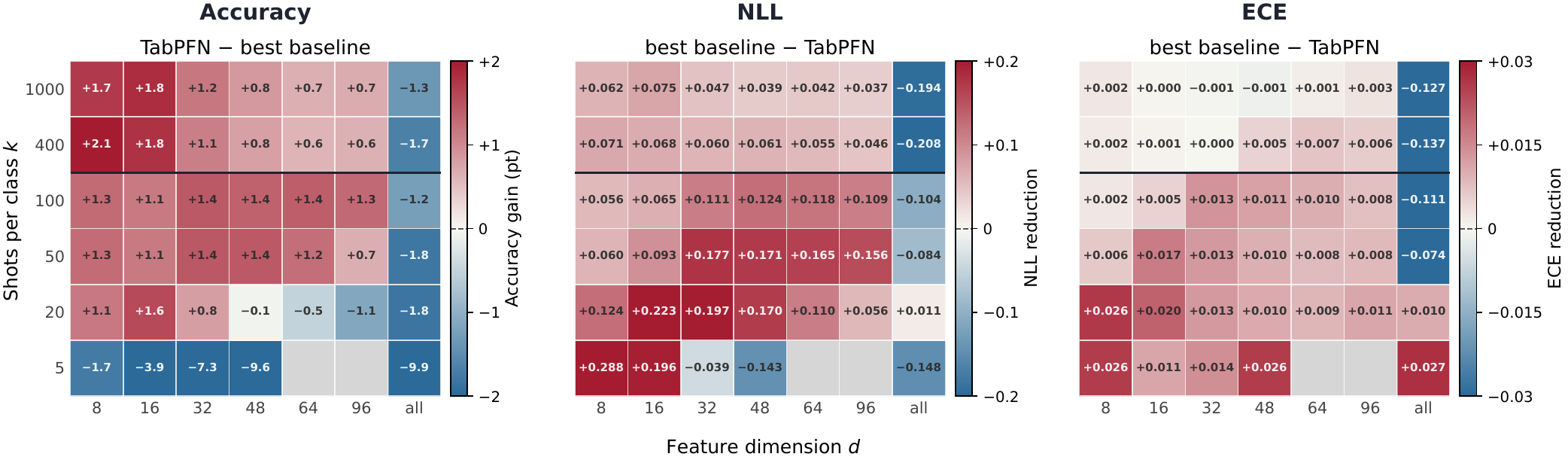}
\caption{%
Mean advantage of TabPFN over the strongest baseline in each $(k, d)$ cell at $C=10$ across Accuracy, NLL, and ECE (26-pair subset $\mathcal{S}_{26}$ for $k \le 100$ below the black horizontal line; $\mathcal{S}_{11}$ for $k \in \{400, 1000\}$ above the black horizontal line). For grey cells ($k = 5$, $d \in \{64, 96\}$), with only $kC = 50$ training examples, PCA yields at most 49 components, so these settings are identical to the $d = \text{all}$ column.%
}
\label{fig:appendix_fixed_grid_cell_means}
\end{figure}

Figure~\ref{fig:appendix_fixed_grid_winrate} plots TabPFN's accuracy win rate against the best of the eight baselines in each episode and against logistic regression (LR) alone, with shaded areas indicating 95\% bootstrap confidence intervals over unique dataset--encoder pairs. The $\mathcal{S}_{11}$ trajectory closely matches $\mathcal{S}_{26}$, confirming that changing task composition does not change the boundary. Figure~\ref{fig:appendix_fixed_grid_cell_means} compares the mean performance of TabPFN against the strongest baseline in each $(k, d)$ cell across Accuracy, NLL, and ECE. Across both fixed subsets, TabPFN achieves lower mean NLL and ECE in nearly every cell with $k \ge 20$ and $d \le 96$, whereas its mean accuracy advantage concentrates at $k \ge 50$ ($d \le 96$) and $k=20$ ($d \le 32$).

\subsection{Appendix F: NLL and ASS for Adapted Backbones}

\label{sec:appendix_f}

\begin{figure}[htbp]
\centering
\includegraphics[width=1.0\textwidth]{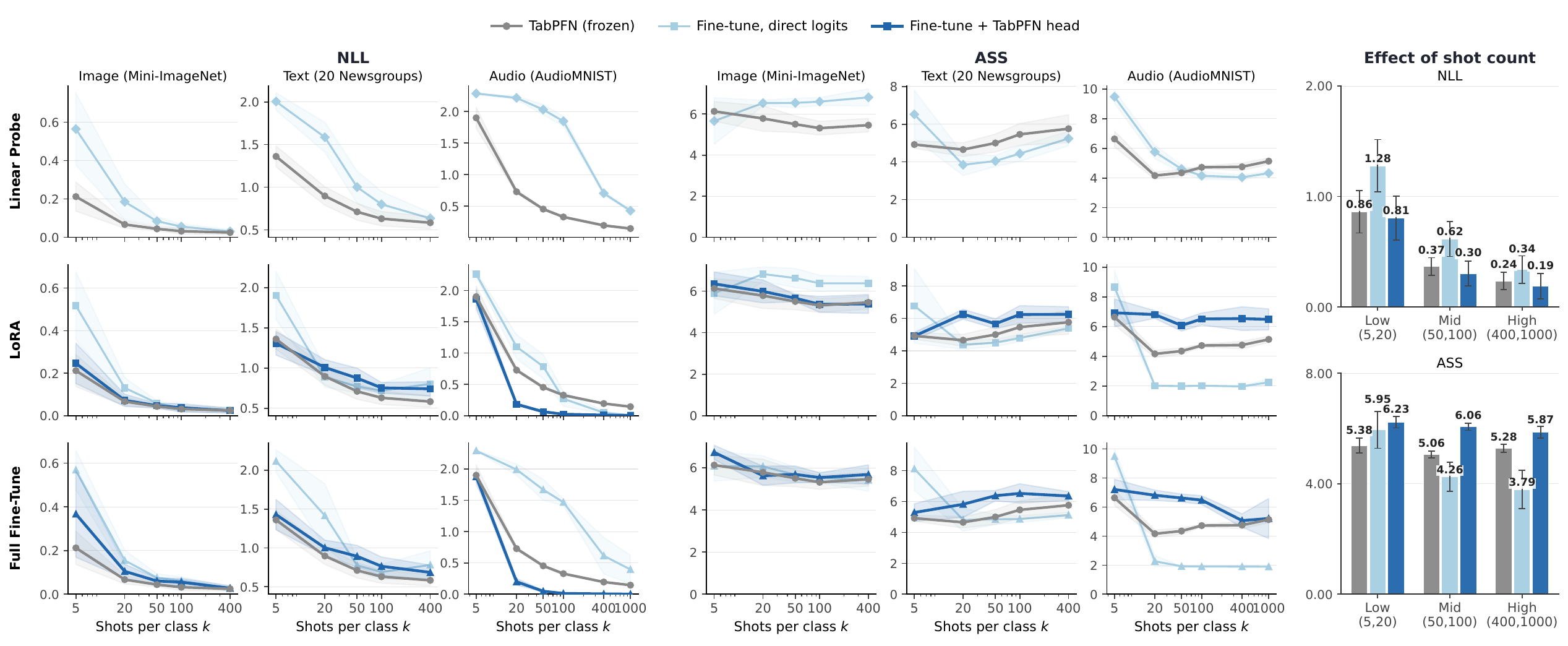}
\caption{%
NLL and ASS for adapted backbones across shot counts.
\emph{Left block}: NLL versus shots per class $k$; \emph{middle block}: ASS versus shots per class $k$; \emph{right block}: aggregate results for the same low-, mid-, and high-shot regimes as Figure~\ref{fig:exp3}. Rows show the fine-tuning method (Linear Probe, LoRA, Full Fine-Tune), and columns in the line-plot blocks show modality and dataset (Image / Mini-ImageNet, Text / 20 Newsgroups, Audio / AudioMNIST). Grey = TabPFN on frozen encoder (reference); light blue = fine-tune with direct linear head; dark blue = fine-tune $+$ TabPFN head. Shading: mean $\pm$1 std over 5 seeds. Right-bar error bars show SEM across configurations within each shot-count bucket.%
}
\label{fig:exp3_appendix_nll_ass}
\end{figure}

Figure~\ref{fig:exp3_appendix_nll_ass} complements the Accuracy and ECE panels of Figure~\ref{fig:exp3} by reporting NLL and ASS for the same backbones, modalities, and shots per class $k$. The NLL results mirror the ECE patterns: the direct adapted heads (light blue) generally yield higher NLL, whereas replacing their linear heads with TabPFN (dark blue) reduces NLL toward the frozen-TabPFN reference (grey). The right summary panels show the same trend after grouping configurations by shot-count regime: the direct heads retain higher NLL from the low- to high-shot regimes, whereas the TabPFN heads reduce NLL to the frozen-TabPFN level or below. Conversely, ASS does not exhibit a uniform ranking across the adaptation methods, reflecting differences in conformal prediction-set efficiency.

\begin{figure}[htbp]
\centering
\includegraphics[width=0.98\textwidth]{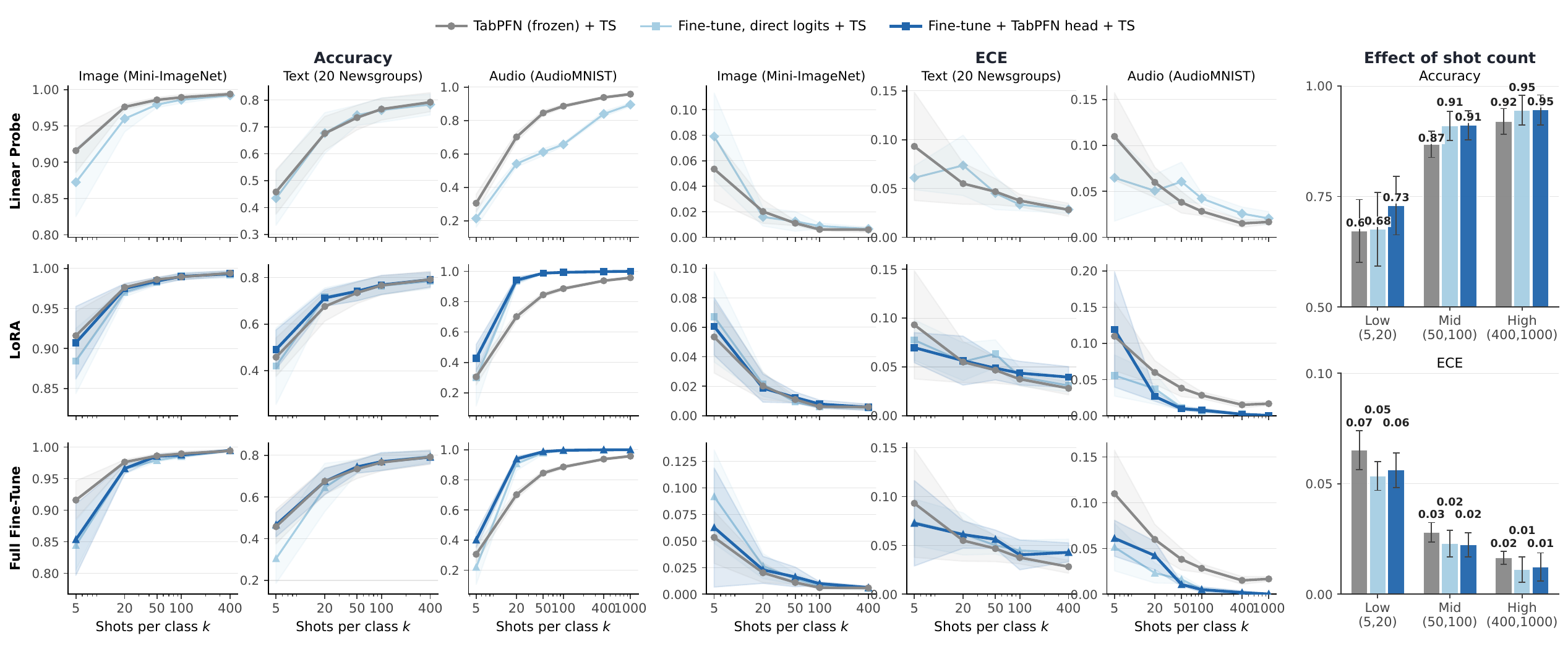}
\caption{%
Accuracy and ECE after backbone adaptation, with every head temperature-scaled ($C=10$, 5 seeds; TabPFN on $d=64$ PCA features). Each head fits a single temperature value on the $20\%$ validation split; the direct head is trained on the remaining $80\%$ of the training examples. All metrics are computed on the held-out test set.%
}
\label{fig:appendix_exp3_ts_acc_ece}
\end{figure}

\begin{figure}[htbp]
\centering
\includegraphics[width=0.98\textwidth]{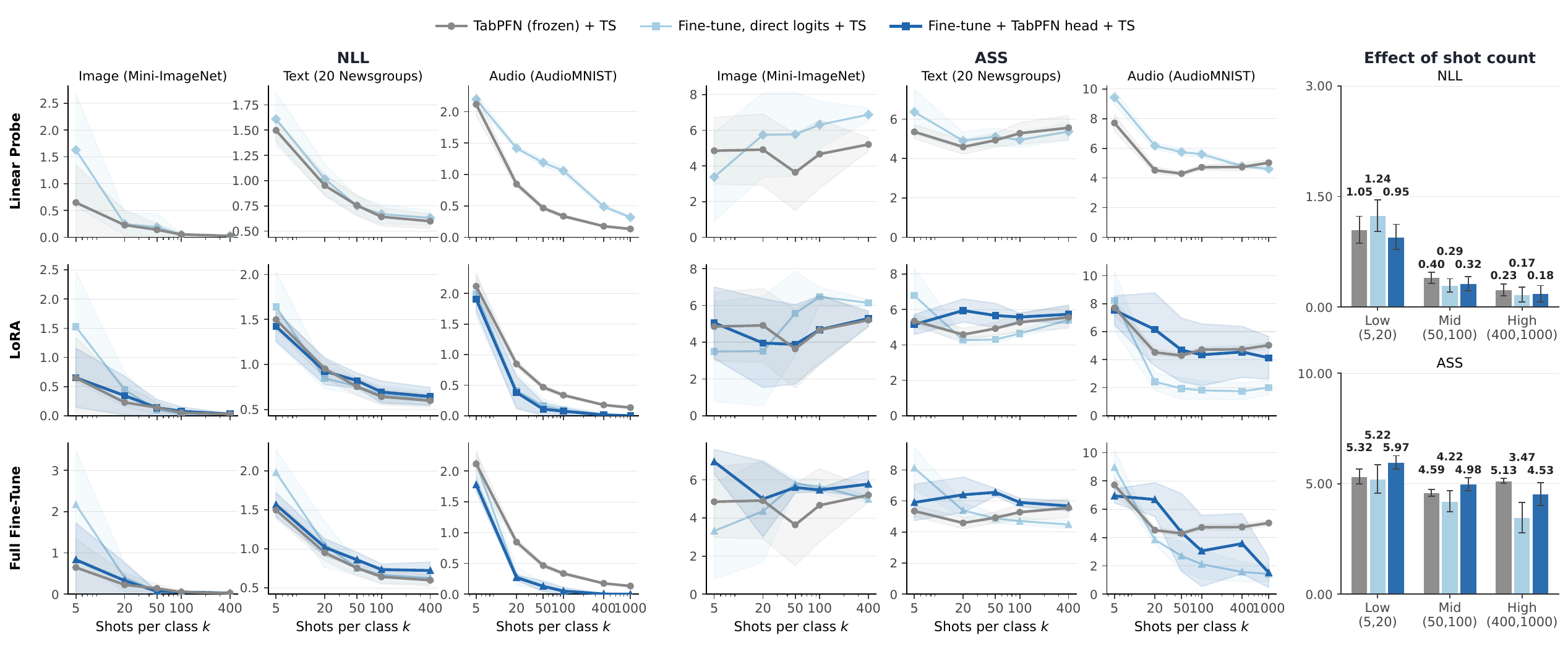}
\caption{%
NLL and ASS after backbone adaptation, with every head temperature-scaled ($C=10$, 5 seeds; TabPFN on $d=64$ PCA features). Each head fits a single temperature value on the $20\%$ validation split; the direct head is trained on the remaining $80\%$ of the training examples. All metrics are computed on the held-out test set.%
}
\label{fig:appendix_exp3_ts_nll_ass}
\end{figure}

In Figure~\ref{fig:exp3}, the direct fine-tuned head is trained on $80\%$ of $\mathcal{D}_{\mathrm{tr}}$ with early stopping on the remaining $20\%$, whereas TabPFN uses all $100\%$ of $\mathcal{D}_{\mathrm{tr}}$ as context. Figures~\ref{fig:appendix_exp3_ts_acc_ece} and~\ref{fig:appendix_exp3_ts_nll_ass} repeat Experiment~3 when every head is temperature-scaled (TS) on the $20\%$ validation split and TabPFN uses only the remaining $80\%$ training split as context. Temperature scaling brings both heads to comparable ECE, with TabPFN\,+\,TS maintaining lower mean NLL on image ($0.134$ vs.\ $0.146$) and audio ($0.103$ vs.\ $0.116$).

\begin{figure}[htbp]
\centering
\includegraphics[width=0.95\textwidth]{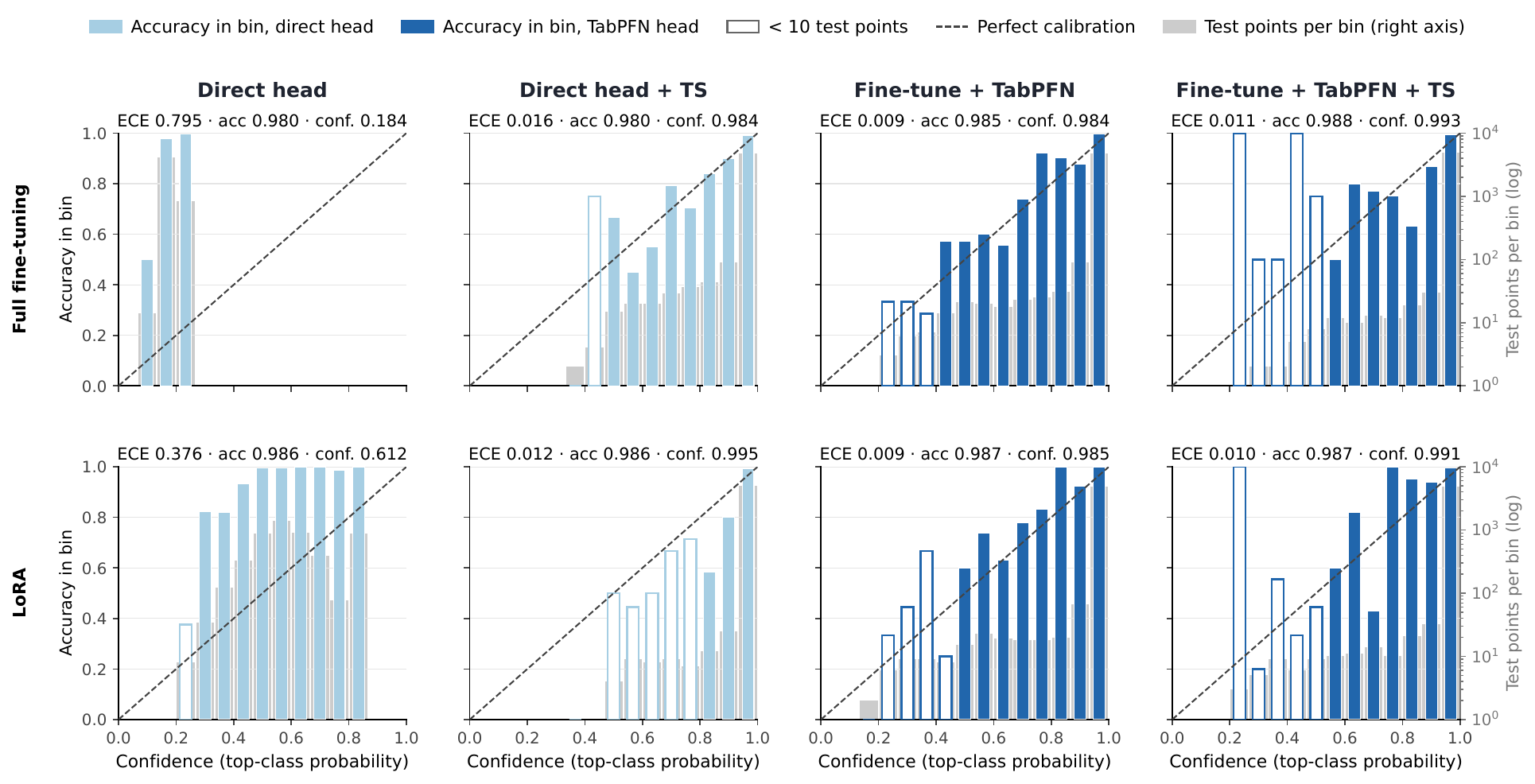}
\caption{%
Reliability diagrams on AudioMNIST (CLAP-HTSAT) at $k=50$, $C=10$ (5 seeds) for full fine-tuning (\emph{top}) and LoRA (\emph{bottom}). Columns compare the uncalibrated direct head, direct head\,+\,TS, fine-tune\,+\,TabPFN, and fine-tune\,+\,TabPFN\,+\,TS, with temperature fitted on the $20\%$ validation split.%
}
\label{fig:appendix_audiomnist_rel}
\end{figure}

\paragraph{AudioMNIST underconfidence diagnosis.}
In Experiment~3, the directly fine-tuned CLAP-HTSAT head on AudioMNIST reaches $\approx 98\%$ accuracy at $k=50$ but has an ECE near $0.80$. The direct head is underconfident because its logits stay small. The CLAP features are L2-normalized, so the logit scale is set by the head weights alone, and the small head learning rate of full fine-tuning lets these weights grow slowly. Accuracy saturates while the logits are still small, and since early stopping monitors validation accuracy rather than loss, training ends at that early checkpoint before the confidence can catch up. Figure~\ref{fig:appendix_audiomnist_rel} shows the resulting reliability diagrams at $k=50$: every prediction from the uncalibrated direct head has confidence $\le 0.25$ despite $0.98$ accuracy ($\text{ECE } 0.795$). Fitting a temperature value on the $20\%$ validation split sharpens these probabilities and reduces ECE to $0.016$ (and $0.376 \to 0.012$ for LoRA), whereas replacing the head with uncalibrated TabPFN achieves an ECE of $0.009$ directly without any calibration step.

\subsection{Appendix G: Extended Top-$k$ Calibration Analysis}
\label{sec:appendix_g}

\begin{figure}[htbp]
\centering
\includegraphics[width=0.98\textwidth]{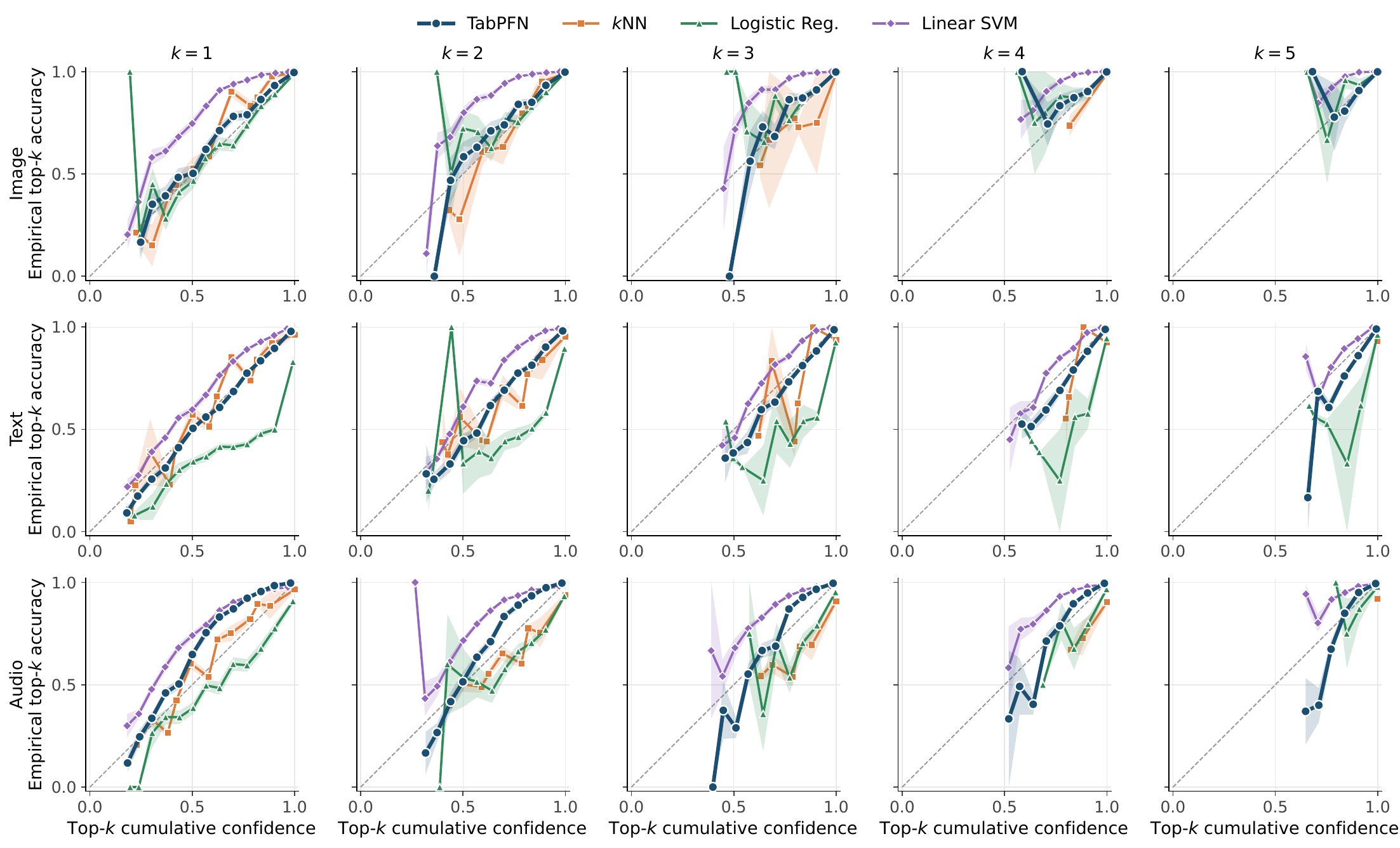}
\caption{%
Top-$k$ reliability diagrams at $k\in\{1,\ldots,5\}$ for the canonical setting (mean $\pm$1 SEM over five seeds per modality). Rows: modality; columns: Top-$k$ rank. Dashed diagonal: perfect calibration.
}
\label{fig:ablation_topkp_reliability}
\end{figure}

This appendix supplements Ablation~2. We hold the canonical setting ($C=10$, $k=100$, $d=96$) fixed and evaluate the four primary classification heads across the valid dataset--encoder pairs, yielding 560 evaluation episodes. Table~\ref{tab:ablation_topkp_ece} reports the Top-$k$ ECE values for $k \in \{1, \ldots, 5\}$; Table~\ref{tab:ablation_topkp_ranks} details the within-episode mean ranks for Top-5 ECE versus ASS over 140 complete episodes (where lower is better). Figure~\ref{fig:ablation_topkp_reliability} presents Top-$k$ reliability diagrams at $k\in\{1,\ldots,5\}$, aggregated over all datasets and encoders within the image, text, and audio modalities.

\begin{table}[ht]
\centering
\caption{Top-$k$ ECE at the canonical setting (mean $\pm$ SEM over 560 ablation episodes). Bold = best per column.}
\label{tab:ablation_topkp_ece}
\begin{adjustbox}{max width=\columnwidth}
\footnotesize
\begin{tabular}{lccccc}
\hline
\textbf{Head} & $k=1$ & $k=2$ & $k=3$ & $k=4$ & $k=5$ \\
\hline
TabPFN        & \textbf{0.031 $\pm$ 0.003} & \textbf{0.019 $\pm$ 0.002} & \textbf{0.012 $\pm$ 0.001} & \textbf{0.008 $\pm$ 0.001} & \textbf{0.005 $\pm$ 0.001} \\
$k$NN         & 0.037 $\pm$ 0.003 & 0.037 $\pm$ 0.004 & 0.045 $\pm$ 0.005 & 0.046 $\pm$ 0.005 & 0.038 $\pm$ 0.004 \\
Logistic Reg. & 0.069 $\pm$ 0.008 & 0.043 $\pm$ 0.005 & 0.028 $\pm$ 0.004 & 0.019 $\pm$ 0.002 & 0.013 $\pm$ 0.002 \\
Linear SVM    & 0.089 $\pm$ 0.003 & 0.072 $\pm$ 0.003 & 0.055 $\pm$ 0.002 & 0.041 $\pm$ 0.002 & 0.029 $\pm$ 0.001 \\
\hline
\end{tabular}
\end{adjustbox}
\end{table}

\begin{table}[ht]
\centering
\caption{Mean rank on Top-5 ECE versus ASS. TabPFN achieves the best Top-5 calibration but trails on ASS; $k$NN shows the opposite pattern.}
\label{tab:ablation_topkp_ranks}
\begin{adjustbox}{max width=\columnwidth}
\footnotesize
\begin{tabular}{lcc}
\hline
\textbf{Head} & \textbf{Top-5 ECE rank $\downarrow$} & \textbf{ASS rank $\downarrow$} \\
\hline
TabPFN        & \textbf{1.48 $\pm$ 0.04} & 2.82 $\pm$ 0.06 \\
$k$NN         & 3.16 $\pm$ 0.08 & \textbf{1.00 $\pm$ 0.00} \\
Logistic Reg. & 1.91 $\pm$ 0.06 & 3.58 $\pm$ 0.05 \\
Linear SVM    & 3.46 $\pm$ 0.06 & 2.60 $\pm$ 0.06 \\
\hline
\end{tabular}
\end{adjustbox}
\end{table}

The APS in the main text is deterministic: it always includes the full boundary class whose cumulative probability crosses the threshold $\hat{q}$. For smooth, well-calibrated distributions like TabPFN's, always keeping the boundary class leads to substantial over-coverage ($\approx 99.8\%$) and enlarges the average prediction-set size (ASS). To compare set efficiency at the intended $90\%$ coverage level, we also evaluate randomized APS~\citep{romano2020classification}, which includes the boundary class probabilistically.

Table~\ref{tab:appendix_aps_randomized} reports empirical coverage and ASS for both APS variants across all nine heads at $C=10$ and $d=96$. Randomized APS brings every method to $89.8\%$--$91.4\%$ coverage and reduces TabPFN's mean set size at $k=100$ from $5.51$ to $1.39$, matching logistic regression ($1.39$) and MLP ($1.40$) for the smallest sets and outperforming $k$NN ($1.50$).

\begin{table}[htbp]
\centering
\caption{
Empirical coverage (\%) and Average Prediction-Set Size (ASS) under deterministic APS and randomized APS at $C=10$, $d=96$, targeting $1-\alpha = 90\%$ coverage. Evaluations cover the 26-pair subset $\mathcal{S}_{26}$ at $k=100$ and the 11-pair subset $\mathcal{S}_{11}$ at $k \in \{400, 1000\}$ (5 seeds averaged within each pair). Each cell reports $\text{Coverage (\%)} / \text{ASS}$. Bold indicates the smallest randomized ASS per shot count.%
}
\label{tab:appendix_aps_randomized}
\begin{adjustbox}{max width=\textwidth}
\footnotesize
\setlength{\tabcolsep}{3pt}
\begin{tabular}{l|cc|cc|cc}
\hline
 & \multicolumn{2}{c|}{\textbf{$k=100$}} & \multicolumn{2}{c|}{\textbf{$k=400$}} & \multicolumn{2}{c}{\textbf{$k=1000$}} \\
\textbf{Head} & Deterministic & Randomised & Deterministic & Randomised & Deterministic & Randomised \\
\hline
TabPFN & 99.8 / 5.51 & 91.4 / \textbf{1.39} & 100.0 / 5.35 & 91.4 / 1.07 & 100.0 / 5.29 & 91.3 / 1.05 \\
$k$NN & 95.5 / 1.69 & 89.8 / 1.50 & 98.3 / 1.36 & 91.3 / 1.18 & 98.6 / 1.28 & 91.3 / 1.12 \\
Logistic Reg. & 99.6 / 6.03 & 90.9 / \textbf{1.39} & 100.0 / 5.81 & 91.0 / 1.01 & 100.0 / 5.54 & 91.1 / 1.01 \\
Linear SVM & 99.6 / 5.31 & 90.9 / 1.65 & 100.0 / 6.00 & 91.1 / 1.19 & 100.0 / 7.07 & 91.1 / 1.14 \\
Random Forest & 98.3 / 4.15 & 90.9 / 1.97 & 99.7 / 4.02 & 90.9 / 1.45 & 99.8 / 3.86 & 91.0 / 1.40 \\
XGBoost & 99.5 / 4.87 & 91.0 / 1.90 & 100.0 / 4.46 & 91.4 / 1.27 & 100.0 / 4.29 & 91.4 / 1.18 \\
LightGBM & 99.7 / 5.61 & 90.8 / 1.60 & 100.0 / 5.15 & 91.0 / 1.08 & 100.0 / 4.78 & 91.0 / 1.06 \\
CatBoost & 99.1 / 4.69 & 91.0 / 1.87 & 99.9 / 4.31 & 91.0 / 1.36 & 100.0 / 4.18 & 91.0 / 1.28 \\
MLP & 99.7 / 5.46 & 90.9 / 1.40 & 100.0 / 5.54 & 91.2 / \textbf{1.00} & 99.9 / 5.14 & 91.0 / \textbf{0.97} \\
\hline
\end{tabular}
\end{adjustbox}
\end{table}

\subsection{Appendix H: ICL Model Comparison and Top-$k$ Calibration}
\label{sec:appendix_h}

\begin{figure}[htbp]
\centering
\includegraphics[width=0.65\textwidth]{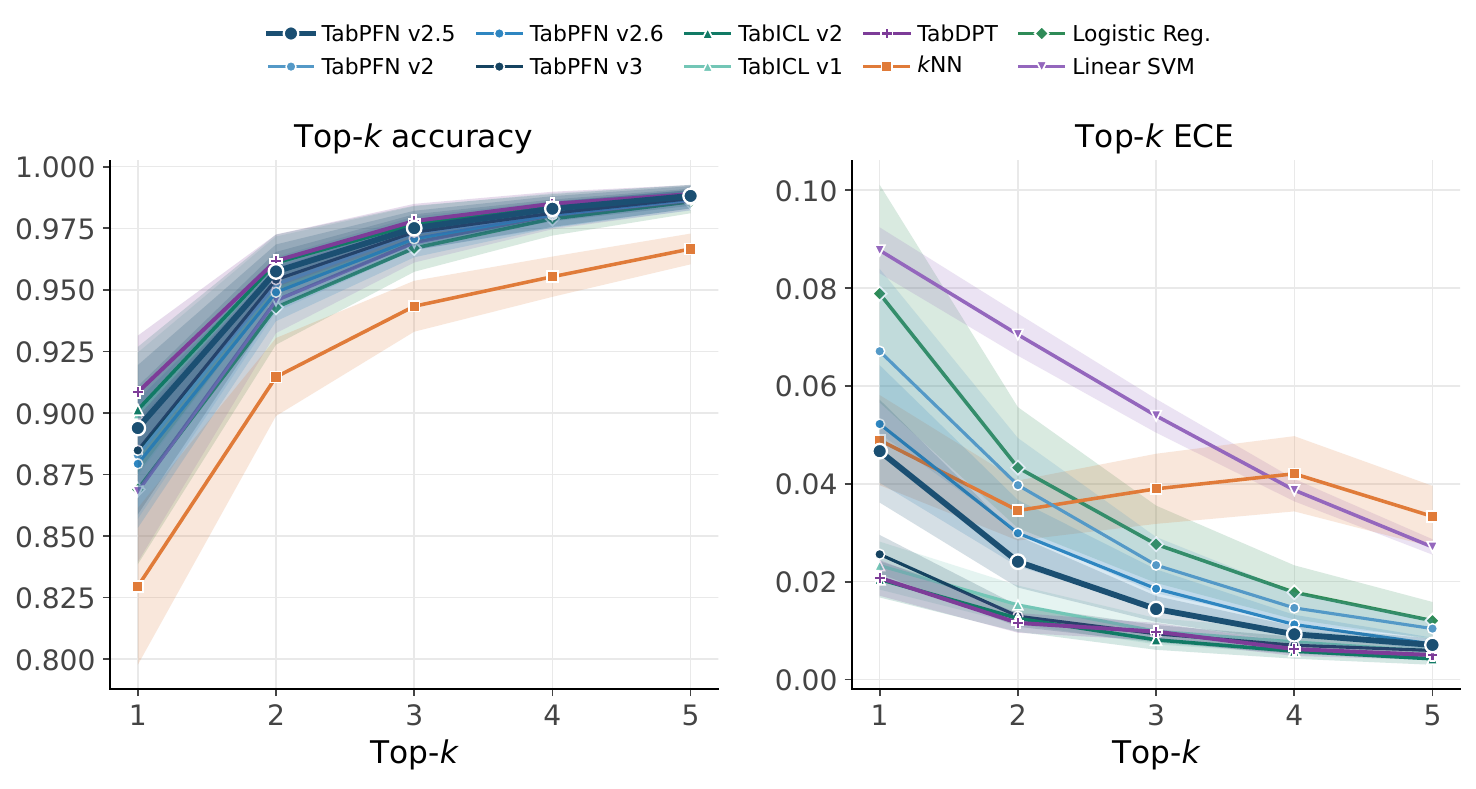}
\caption{%
Top-$k$ calibration for seven ICL models and three classical heads at the canonical setting ($C=10$, $k=100$, $d=96$), mean $\pm$1 SEM over $1{,}400$ ablation episodes. \emph{Left:} Top-$k$ accuracy. \emph{Right:} Top-$k$ ECE.%
}
\label{fig:ablation5_topkp}
\end{figure}

This appendix supplements Ablation~3. The primary question is whether ICL model heads transfer calibration to frozen multimodal embeddings. We therefore evaluate seven ICL foundation models and three classical supervised heads at the canonical setting ($C=10$, $k=100$, $d=96$). For each ICL model, we use its publicly released classifier checkpoint and the default inference settings provided by its corresponding implementation. Figure~\ref{fig:ablation5_topkp} shows Top-$k$ accuracy and Top-$k$ ECE for $k\in\{1,\ldots,5\}$.

\subsection{Appendix I: Comparison with Temperature-Scaled Baselines}
\label{sec:appendix_i}

The main text evaluates classification heads using its standard probability-estimation procedure, without applying additional post-hoc recalibration. To test whether a standard post-hoc calibrator closes the reliability gap when sufficient labeled examples are available, we apply temperature scaling (TS)~\citep{guo2017calibration} to all eight baseline heads as well as TabPFN. 

We evaluate the canonical 10-class setting ($C=10$, $d=96$) across moderate-to-high shot counts $k \in \{100, 400, 1000\}$, covering 26, 19, and 11 unique dataset--encoder pairs, respectively, over 5 random seeds (averaging across seeds within each pair before aggregating, with 95\% confidence intervals computed via 10000 bootstrap resamples over pairs). For each head and episode, a single scalar temperature $T$ is fitted by minimizing cross-entropy loss on 5-fold cross-validation log-probabilities from the $k \cdot C$ training examples, after which the classification head is refit on the full training dataset. TabPFN\,+\,TS follows the exact same 5-fold procedure.

\begin{figure}[htbp]
\centering
\includegraphics[width=0.95\textwidth]{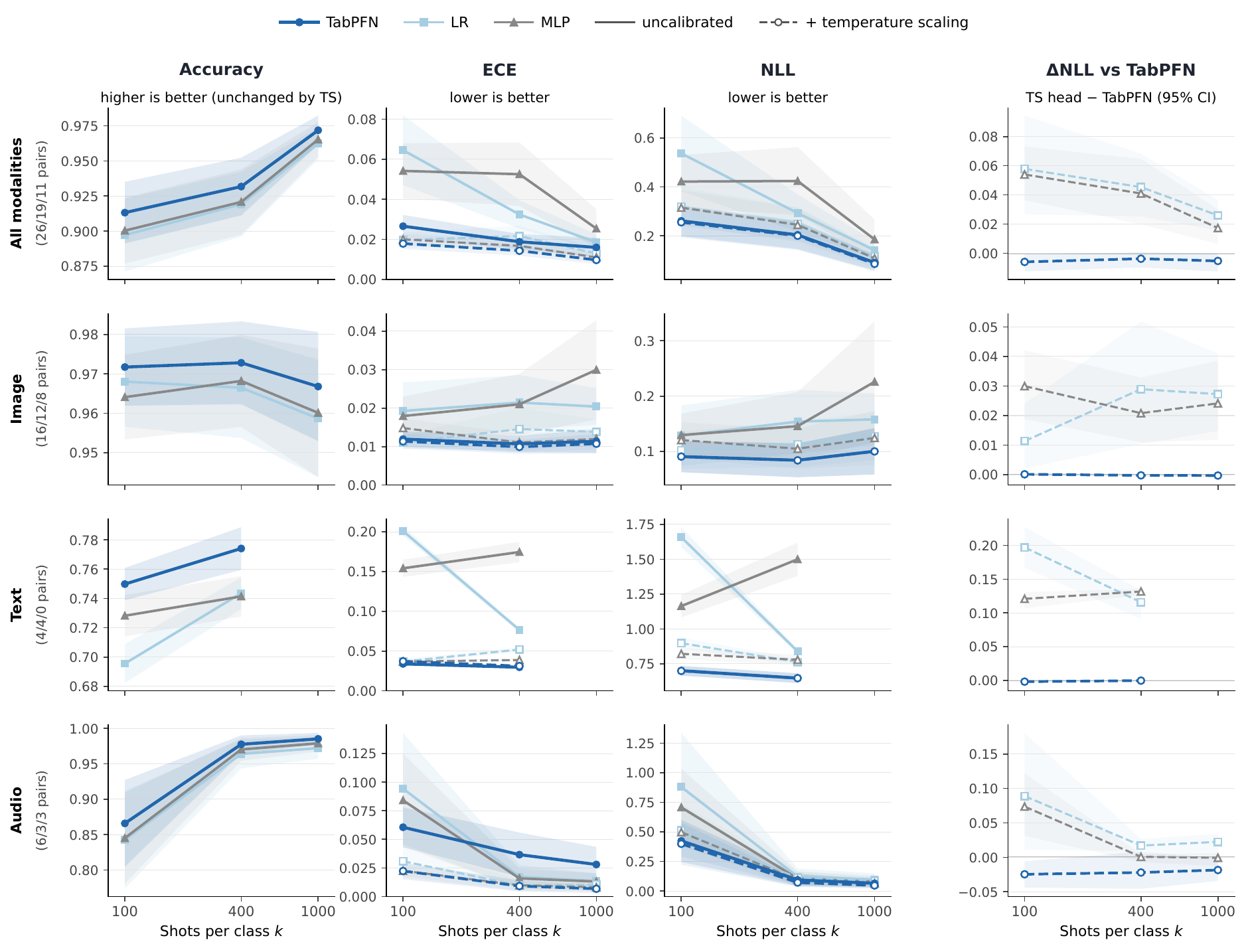}
\caption{%
Comparison of TabPFN against logistic regression (LR) and MLP with and without temperature scaling (TS) at $C=10$, $d=96$, and $k \in \{100, 400, 1000\}$. The first three columns are mean accuracy, ECE, and NLL across unique dataset--encoder pairs (shaded regions indicate $\pm$1 SEM; TS accuracy curves are omitted because TS does not change accuracy). The fourth column reports the paired difference $\Delta\text{NLL}$ ($\text{head + TS} - \text{uncalibrated TabPFN}$); the shaded band is its 95\% confidence interval, computed by resampling dataset--encoder pairs. Values above zero favour uncalibrated TabPFN.
}
\label{fig:appendix_ts_baselines}
\end{figure}

Figure~\ref{fig:appendix_ts_baselines} compares TabPFN against logistic regression (LR) and MLP, with and without TS across modalities. While TS brings the ECE of LR and MLP down to the level of uncalibrated TabPFN, their NLL remains consistently higher. Table~\ref{tab:appendix_ts_all} reports mean ECE, NLL, and accuracy across all nine heads before and after temperature scaling, and Table~\ref{tab:appendix_ts_paired} provides the paired differences of 8 baseline heads + TS against both uncalibrated TabPFN and TabPFN\,+\,TS with 95\% bootstrap confidence intervals.

\begin{table}[htbp]
\centering
\caption{Mean ECE, NLL, and top-1 accuracy for all nine classification heads with and without temperature scaling (TS) at $C=10$, $d=96$, evaluated over 26, 19, and 11 unique dataset--encoder pairs at $k=100$, $400$, and $1000$ (5 seeds averaged within each pair). Lower is better for ECE and NLL; higher is better for Accuracy. Bold indicates the best mean value per column.%
}
\label{tab:appendix_ts_all}
\begin{adjustbox}{max width=\textwidth}
\footnotesize
\setlength{\tabcolsep}{3pt}
\begin{tabular}{l|ccc|ccc|ccc}
\hline
& \multicolumn{3}{c|}{\textbf{ECE}} & \multicolumn{3}{c|}{\textbf{NLL}} & \multicolumn{3}{c}{\textbf{Acc}} \\
\textbf{Head} & $k=100$ & $k=400$ & $k=1000$ & $k=100$ & $k=400$ & $k=1000$ & $k=100$ & $k=400$ & $k=1000$ \\
\hline
TabPFN & 0.027 & 0.019 & 0.016 & 0.261 & 0.204 & 0.091 & \textbf{0.913} & \textbf{0.932} & \textbf{0.972} \\
TabPFN + TS & \textbf{0.018} & \textbf{0.014} & \textbf{0.010} & \textbf{0.256} & \textbf{0.200} & \textbf{0.086} & \textbf{0.913} & \textbf{0.932} & \textbf{0.972} \\
\hline
$k$NN & 0.034 & 0.030 & 0.023 & 1.530 & 0.980 & 0.496 & 0.866 & 0.911 & 0.950 \\
$k$NN + TS & 0.037 & 0.028 & 0.017 & 1.521 & 0.973 & 0.488 & 0.866 & 0.911 & 0.950 \\
Logistic Reg. & 0.065 & 0.032 & 0.019 & 0.537 & 0.293 & 0.142 & 0.897 & 0.919 & 0.962 \\
Logistic Reg. + TS & 0.020 & 0.022 & 0.013 & 0.319 & 0.250 & 0.117 & 0.897 & 0.919 & 0.962 \\
Linear SVM & 0.086 & 0.039 & 0.022 & 0.370 & 0.274 & 0.128 & 0.895 & 0.912 & 0.961 \\
Linear SVM + TS & 0.020 & 0.017 & 0.012 & 0.309 & 0.257 & 0.119 & 0.899 & 0.915 & 0.962 \\
Random Forest & 0.256 & 0.197 & 0.182 & 0.704 & 0.507 & 0.391 & 0.875 & 0.906 & 0.944 \\
Random Forest + TS & 0.021 & 0.017 & 0.013 & 0.411 & 0.290 & 0.172 & 0.875 & 0.906 & 0.944 \\
XGBoost & 0.036 & 0.030 & 0.027 & 0.433 & 0.296 & 0.171 & 0.864 & 0.906 & 0.949 \\
XGBoost + TS & 0.025 & 0.018 & 0.014 & 0.424 & 0.288 & 0.158 & 0.864 & 0.906 & 0.949 \\
LightGBM & 0.058 & 0.044 & 0.023 & 0.459 & 0.332 & 0.166 & 0.878 & 0.914 & 0.956 \\
LightGBM + TS & 0.022 & 0.017 & 0.013 & 0.386 & 0.268 & 0.138 & 0.878 & 0.914 & 0.956 \\
CatBoost & 0.125 & 0.077 & 0.074 & 0.492 & 0.351 & 0.244 & 0.873 & 0.902 & 0.940 \\
CatBoost + TS & 0.023 & 0.018 & 0.015 & 0.377 & 0.290 & 0.181 & 0.873 & 0.902 & 0.940 \\
MLP & 0.054 & 0.053 & 0.025 & 0.422 & 0.424 & 0.186 & 0.900 & 0.921 & 0.965 \\
MLP + TS & 0.020 & 0.017 & 0.011 & 0.315 & 0.245 & 0.108 & 0.900 & 0.921 & 0.965 \\
\hline
\end{tabular}
\end{adjustbox}
\end{table}

\begin{table}[htbp]
\centering
\caption{
Paired differences ($\text{TS baseline} - \text{reference}$) averaged across unique dataset--encoder pairs at $C=10$, $d=96$, $k \in \{100, 400, 1000\}$, with 95\% confidence intervals computed via 10000 bootstrap resamples over pairs. The left block compares each temperature-scaled baseline against uncalibrated TabPFN, and the right block compares against TabPFN\,+\,TS. Positive $\Delta\text{NLL}$ and $\Delta\text{ECE}$ values and negative $\Delta\text{Acc}$ values (in percentage points, pt) favor the TabPFN reference.%
}
\label{tab:appendix_ts_paired}
\begin{adjustbox}{max width=\textwidth}
\footnotesize
\setlength{\tabcolsep}{3pt}
\begin{tabular}{cl|cccc|cc}
\hline
& & \multicolumn{4}{c|}{\textbf{vs.\ TabPFN}} & \multicolumn{2}{c}{\textbf{vs.\ TabPFN + TS}} \\
$k$ & \textbf{TS baseline} & $\Delta$NLL & TabPFN lower NLL & $\Delta$ECE & $\Delta$Acc (pt) & $\Delta$NLL & $\Delta$ECE \\
\hline
100 & $k$NN + TS & +1.260 [+0.658, +1.993] & 25/26 & +0.010 [+0.002, +0.018] & $-$4.7 [$-$7.9, $-$2.0] & +1.266 [+0.652, +2.025] & +0.019 [+0.010, +0.029] \\
 & Logistic Reg. + TS & +0.058 [+0.027, +0.094] & 19/26 & $-$0.007 [$-$0.015, $-$0.001] & $-$1.6 [$-$2.4, $-$0.8] & +0.064 [+0.031, +0.101] & +0.002 [$-$0.001, +0.005] \\
 & Linear SVM + TS & +0.047 [+0.026, +0.071] & 24/26 & $-$0.007 [$-$0.015, +0.000] & $-$1.4 [$-$2.0, $-$0.9] & +0.053 [+0.030, +0.079] & +0.002 [+0.001, +0.003] \\
 & Random Forest + TS & +0.149 [+0.089, +0.216] & 26/26 & $-$0.006 [$-$0.013, +0.000] & $-$3.9 [$-$6.1, $-$2.0] & +0.155 [+0.092, +0.226] & +0.003 [+0.001, +0.005] \\
 & XGBoost + TS & +0.163 [+0.109, +0.224] & 26/26 & $-$0.001 [$-$0.010, +0.006] & $-$4.9 [$-$7.3, $-$3.0] & +0.169 [+0.112, +0.233] & +0.007 [+0.005, +0.010] \\
 & LightGBM + TS & +0.124 [+0.085, +0.168] & 26/26 & $-$0.004 [$-$0.013, +0.003] & $-$3.5 [$-$5.1, $-$2.2] & +0.130 [+0.087, +0.179] & +0.005 [+0.003, +0.006] \\
 & CatBoost + TS & +0.116 [+0.073, +0.166] & 26/26 & $-$0.004 [$-$0.012, +0.003] & $-$4.0 [$-$6.0, $-$2.3] & +0.122 [+0.074, +0.177] & +0.005 [+0.003, +0.007] \\
 & MLP + TS & +0.054 [+0.036, +0.073] & 25/26 & $-$0.007 [$-$0.016, +0.001] & $-$1.3 [$-$1.7, $-$0.9] & +0.060 [+0.041, +0.082] & +0.002 [+0.000, +0.005] \\
\hline
400 & $k$NN + TS & +0.769 [+0.411, +1.162] & 19/19 & +0.009 [+0.001, +0.018] & $-$2.1 [$-$3.8, $-$0.6] & +0.772 [+0.419, +1.169] & +0.014 [+0.007, +0.022] \\
 & Logistic Reg. + TS & +0.045 [+0.025, +0.068] & 18/19 & +0.003 [$-$0.007, +0.011] & $-$1.3 [$-$1.9, $-$0.7] & +0.049 [+0.028, +0.071] & +0.007 [+0.002, +0.014] \\
 & Linear SVM + TS & +0.053 [+0.028, +0.081] & 18/19 & $-$0.002 [$-$0.008, +0.003] & $-$1.7 [$-$2.5, $-$0.9] & +0.056 [+0.030, +0.085] & +0.003 [+0.001, +0.005] \\
 & Random Forest + TS & +0.086 [+0.048, +0.129] & 19/19 & $-$0.001 [$-$0.007, +0.003] & $-$2.6 [$-$4.2, $-$1.2] & +0.089 [+0.048, +0.139] & +0.003 [+0.002, +0.005] \\
 & XGBoost + TS & +0.084 [+0.053, +0.120] & 19/19 & $-$0.001 [$-$0.007, +0.003] & $-$2.5 [$-$3.9, $-$1.4] & +0.087 [+0.053, +0.129] & +0.004 [+0.002, +0.005] \\
 & LightGBM + TS & +0.063 [+0.043, +0.088] & 19/19 & $-$0.002 [$-$0.008, +0.003] & $-$1.8 [$-$2.8, $-$1.0] & +0.067 [+0.043, +0.096] & +0.003 [+0.002, +0.004] \\
 & CatBoost + TS & +0.086 [+0.050, +0.125] & 19/19 & $-$0.001 [$-$0.006, +0.002] & $-$3.0 [$-$4.6, $-$1.7] & +0.089 [+0.051, +0.136] & +0.003 [+0.001, +0.005] \\
 & MLP + TS & +0.041 [+0.020, +0.065] & 18/19 & $-$0.002 [$-$0.009, +0.003] & $-$1.1 [$-$1.7, $-$0.6] & +0.045 [+0.025, +0.068] & +0.002 [+0.001, +0.004] \\
\hline
1000 & $k$NN + TS & +0.397 [+0.181, +0.662] & 10/11 & +0.001 [$-$0.007, +0.007] & $-$2.2 [$-$4.0, $-$0.7] & +0.402 [+0.187, +0.671] & +0.007 [+0.003, +0.013] \\
 & Logistic Reg. + TS & +0.026 [+0.015, +0.036] & 11/11 & $-$0.003 [$-$0.011, +0.003] & $-$0.9 [$-$1.4, $-$0.5] & +0.031 [+0.018, +0.044] & +0.003 [+0.002, +0.005] \\
 & Linear SVM + TS & +0.028 [+0.018, +0.039] & 10/11 & $-$0.004 [$-$0.012, +0.001] & $-$1.0 [$-$1.6, $-$0.6] & +0.034 [+0.020, +0.048] & +0.002 [+0.001, +0.003] \\
 & Random Forest + TS & +0.081 [+0.033, +0.137] & 11/11 & $-$0.003 [$-$0.009, +0.002] & $-$2.8 [$-$4.8, $-$1.1] & +0.086 [+0.034, +0.149] & +0.004 [+0.002, +0.006] \\
 & XGBoost + TS & +0.067 [+0.033, +0.106] & 11/11 & $-$0.002 [$-$0.009, +0.002] & $-$2.3 [$-$4.0, $-$0.9] & +0.072 [+0.034, +0.118] & +0.004 [+0.002, +0.006] \\
 & LightGBM + TS & +0.048 [+0.026, +0.072] & 11/11 & $-$0.003 [$-$0.010, +0.001] & $-$1.6 [$-$2.7, $-$0.7] & +0.053 [+0.027, +0.084] & +0.003 [+0.001, +0.005] \\
 & CatBoost + TS & +0.090 [+0.041, +0.148] & 11/11 & $-$0.001 [$-$0.007, +0.003] & $-$3.2 [$-$5.6, $-$1.3] & +0.095 [+0.042, +0.161] & +0.005 [+0.003, +0.008] \\
 & MLP + TS & +0.017 [+0.006, +0.030] & 10/11 & $-$0.005 [$-$0.014, +0.001] & $-$0.7 [$-$1.0, $-$0.3] & +0.023 [+0.013, +0.034] & +0.001 [+0.000, +0.002] \\
\hline
\end{tabular}
\end{adjustbox}
\end{table}

\subsection{Appendix J: Choice of Dimensionality Reduction}
\label{sec:appendix_j}

Our pipeline uses PCA because it is label-free and preserves the maximum variance in a rank-$d$ linear subspace. To test whether non-linear or supervised projections work better with TabPFN, we compare PCA at $d=96$, $k \in \{20, 100\}$, 26 dataset--encoder pairs (3 seeds) against three alternatives: a supervised 96-unit hidden-layer projection from a MLP trained on the context labels, unsupervised Kernel PCA (RBF kernel with median-heuristic bandwidth), and Gaussian random projection, alongside a MLP classifier trained directly on all raw features.

Table~\ref{tab:appendix_dim_reduction} shows that standard PCA achieves the highest accuracy and lowest NLL at both shot counts. Kernel PCA matches PCA on ECE but trails slightly on accuracy and NLL. Table~\ref{tab:appendix_pca_cost} further evaluates how much task-relevant information is lost when compressing full encoder embeddings to $d=96$ across individual datasets. Although PCA compression reduces logistic regression accuracy (LR, $d=96$ vs.\ LR, $d=\texttt{all}$) on several datasets, TabPFN at $d=96$ recovers nearly all of this compression loss, matching all-feature logistic regression in mean accuracy across $k \in \{100, 400, 1000\}$.

\begin{table}[t]
\centering
\caption{Comparison of PCA + TabPFN against alternative dimensionality reduction methods + TabPFN, with $d=96$, $C=10$, and $k \in \{20, 100\}$, evaluated on 26 dataset--encoder pairs, 3 seeds. The last row is a reference of an MLP classifier applied directly to the full standardized embedding with no dimensionality reduction.
}
\label{tab:appendix_dim_reduction}
\begin{adjustbox}{max width=\columnwidth}
\footnotesize
\begin{tabular}{l|cc|cc|cc}
\hline
& \multicolumn{2}{c|}{\textbf{Acc}} & \multicolumn{2}{c|}{\textbf{ECE}} & \multicolumn{2}{c}{\textbf{NLL}} \\
\textbf{Reduction to $d=96$} & $k=20$ & $k=100$ & $k=20$ & $k=100$ & $k=20$ & $k=100$ \\
\hline
PCA (paper) & 0.860 & 0.915 & 0.032 & 0.026 & 0.422 & 0.256 \\
MLP projection (supervised) & 0.852 & 0.910 & 0.051 & 0.040 & 0.520 & 0.352 \\
Kernel PCA (RBF) & 0.842 & 0.908 & 0.033 & 0.025 & 0.472 & 0.274 \\
Gaussian random projection & 0.779 & 0.863 & 0.055 & 0.044 & 0.673 & 0.419 \\
\hline
\textit{MLP classifier on all features} & 0.850 & 0.910 & 0.055 & 0.044 & 0.531 & 0.342 \\
\hline
\end{tabular}
\end{adjustbox}
\end{table}

\begin{table}[t]
\centering
\caption{Comparison of TabPFN ($d=96$) with logistic regression (LR, $d=96$ and $d=\texttt{all}$) at $C=10$ (5 seeds). The upper block breaks down results per dataset at $k=100$ with 26-pair subset ($\mathcal{S}_{26}$), and the lower block reports means over pairs at $k=100$ ($\mathcal{S}_{26}$) and $k \in \{400, 1000\}$ ($\mathcal{S}_{11}$).%
}
\label{tab:appendix_pca_cost}
\begin{adjustbox}{max width=\textwidth}
\footnotesize
\setlength{\tabcolsep}{3pt}
\begin{tabular}{l|ccc|ccc|ccc}
\hline
& \multicolumn{3}{c|}{\textbf{Acc}} & \multicolumn{3}{c|}{\textbf{ECE}} & \multicolumn{3}{c}{\textbf{NLL}} \\
& LR & LR & TabPFN & LR & LR & TabPFN & LR & LR & TabPFN \\
\textbf{Dataset} & $d=96$ & all & $d=96$ & $d=96$ & all & $d=96$ & $d=96$ & all & $d=96$ \\
\hline
20 Newsgroups & 0.696 & 0.732 & 0.749 & 0.201 & 0.134 & 0.034 & 1.659 & 1.037 & 0.701 \\
AudioMNIST & 0.944 & 0.955 & 0.958 & 0.020 & 0.016 & 0.058 & 0.184 & 0.145 & 0.166 \\
CIFAR-10 & 0.918 & 0.932 & 0.929 & 0.049 & 0.027 & 0.021 & 0.352 & 0.226 & 0.220 \\
STL-10 & 0.982 & 0.986 & 0.984 & 0.011 & 0.009 & 0.008 & 0.066 & 0.049 & 0.052 \\
Speech Commands & 0.741 & 0.796 & 0.773 & 0.169 & 0.095 & 0.064 & 1.575 & 0.903 & 0.682 \\
Stanford Dogs & 0.982 & 0.985 & 0.985 & 0.010 & 0.009 & 0.011 & 0.061 & 0.048 & 0.053 \\
mini-ImageNet & 0.990 & 0.991 & 0.989 & 0.007 & 0.007 & 0.007 & 0.035 & 0.033 & 0.038 \\
\hline
\textit{Mean, $k=100$} & 0.897 & 0.914 & 0.913 & 0.065 & 0.041 & 0.027 & 0.537 & 0.335 & 0.261 \\
\textit{Mean, $k=400$} & 0.956 & 0.967 & 0.967 & 0.026 & 0.018 & 0.020 & 0.190 & 0.126 & 0.108 \\
\textit{Mean, $k=1000$} & 0.962 & 0.970 & 0.972 & 0.019 & 0.020 & 0.016 & 0.141 & 0.130 & 0.091 \\
\hline
\end{tabular}
\end{adjustbox}
\end{table}

\end{document}